\pdfoutput=1

\documentclass[11pt]{article}

\usepackage[]{EACL2023}

\usepackage{times}
\usepackage{latexsym}
\usepackage{paralist}
\usepackage{multirow}
\usepackage{booktabs}
\usepackage{graphicx}

\usepackage[T1]{fontenc}

\usepackage[utf8]{inputenc}

\usepackage{microtype}

\usepackage{inconsolata}
\usepackage{hyperref}
\usepackage{breakcites}

\newcommand{\ignore}[1]{}

\newcommand{\Sref}[1]{\S\ref{#1}}
\newcommand{\fref}[1]{Fig.~\ref{#1}}

\newcommand{\tref}[1]{Tab.~\ref{#1}}

\newcommand{\Aref}[1]{Appendix~\ref{#1}}

\title{Language Generation Models Can Cause Harm: \\So What Can We Do About It? An Actionable Survey}

\author{Sachin Kumar$^{*,\clubsuit}$ \quad Vidhisha Balachandran$^{*,\clubsuit}$ \quad Lucille Njoo$^\heartsuit$ \\ \quad \textbf{Antonios Anastasopoulos}$^\diamondsuit$ \quad \textbf{Yulia Tsvetkov}$^\heartsuit$ \\
$^\clubsuit$Language Technologies Institute, Carnegie Mellon University, Pittsburgh PA \\
$^\diamondsuit$Department of Computer Science, George Mason University, Fairfax, VA \\
 $^\heartsuit$Paul G.~Allen School of Computer Science \& Engineering, University of Washington, Seattle WA \\
\texttt{\small \{sachink,vbalacha\}@cs.cmu.edu, lnjoo@cs.washington.edu, antonis@gmu.edu, yuliats@cs.washington.edu}}

\begin{document}
\maketitle
\def\thefootnote{*}\footnotetext{Equal contribution}\def\thefootnote{\arabic{footnote}}
\begin{abstract}
Recent advances in the capacity of large language models to generate human-like text 
have resulted in their increased adoption in user-facing settings. 
In parallel, these improvements have prompted a heated discourse around the risks of societal harms they introduce, whether inadvertent or malicious.
Several studies have explored these harms and called for their mitigation via 
development of safer, fairer models. 
Going beyond enumerating the risks of harms, this work provides a survey of \textit{practical methods} for addressing potential threats and societal harms from language generation models.
We draw on several prior works' taxonomies of language model risks to present a structured overview of strategies 
for detecting and ameliorating different kinds of risks/harms of language generators. 
Bridging diverse strands of research, this survey aims to serve as a practical guide for both LM researchers and practitioners, with explanations of different mitigation strategies' motivations, their limitations, and open problems for future research. 

\end{abstract}

\section{Introduction}
The new wave of large language models \citep[LMs; ][]{NEURIPS2020_1457c0d6,https://doi.org/10.48550/arxiv.2204.02311,https://doi.org/10.48550/arxiv.2205.01068} 
capable of generating text with human-like fluency, coherence, and realism \citep{ zellers2019defending,ippolito-etal-2020-automatic}
has caused a paradigm shift in our society.
\footnote{While the majority of these models are trained on English, recent studies have also obtained similar advancements in other languages~\citep{Lin2021FewshotLW, https://doi.org/10.48550/arxiv.2204.07580}.}  
With applications like \href{https://openai.com/blog/chatgpt/}{OpenAI's ChatGPT}, Microsoft's \href{https://www.bing.com/new}{Bing}, and Google's \href{https://blog.google/technology/ai/bard-google-ai-search-updates/}{Bard}, bringing such LMs directly to users, we are beginning to see the impact in fields like education \cite{chatgpt-education, chatgpt-education2}, healthcare \cite{Patel2023ChatGPTTF}, law \cite{ChatGPT2022TheIO}, science \citep{chatgpt-science}, and more. 
Since language is inherently a tool of power---the primary means by which people and societies perpetuate stereotypes
and manipulate opinions 
\cite[\textit{inter alia}]{bar2013stereotyping,chong2007framing}---LMs that are deployed to millions of users also hold similar power, but
our understanding of their risks/harms has lagged behind \cite{10.1145/3442188.3445922}. 

\begin{figure}
    \centering
    \includegraphics[width=0.4\textwidth]{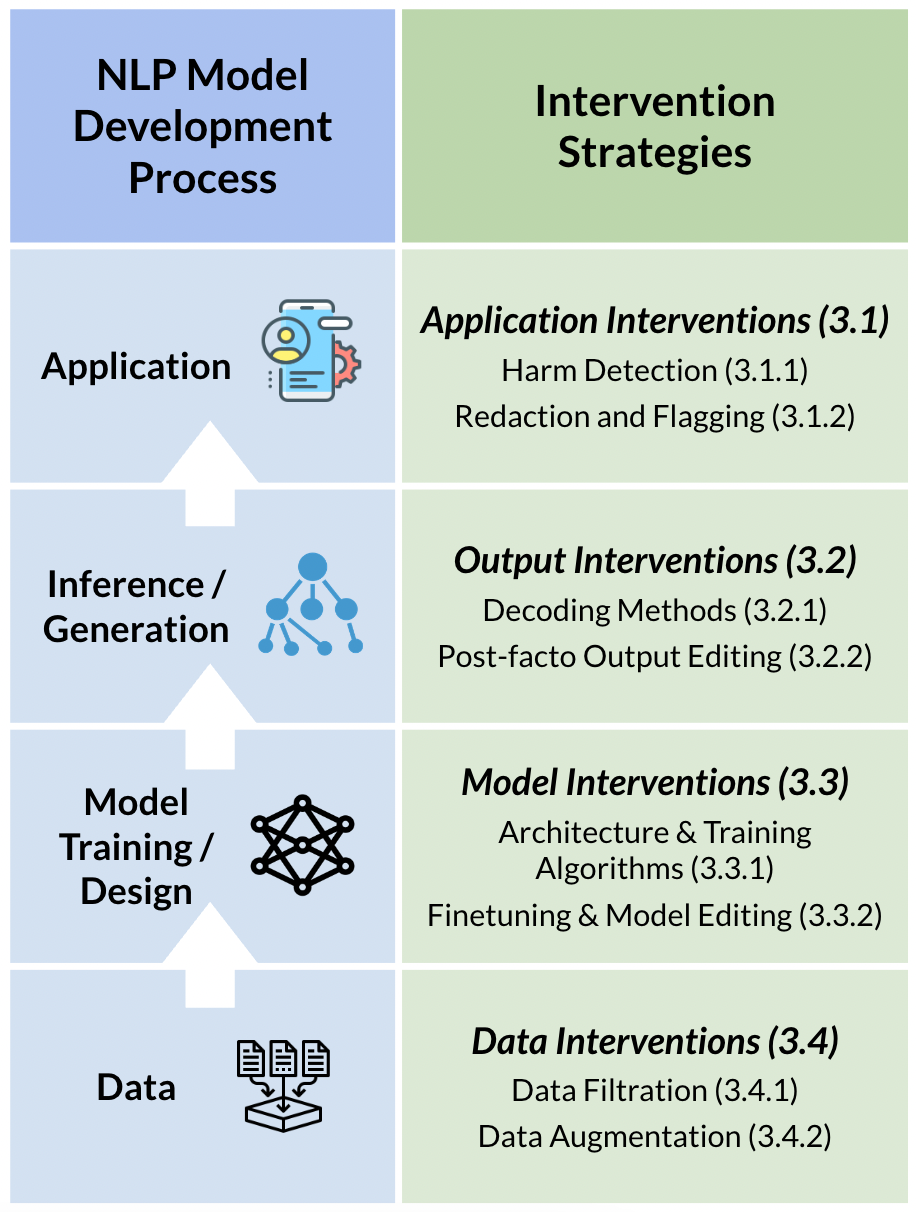}
    \vspace{-.5em}
    \caption{Overview of Intervention Strategies. A typical ML/NLP model development process involves data collection/curation, model training and design, inference, and finally application deployment. For each phase of this development cycle, different techniques can be adopted to mitigate harms. Our survey presents a taxonomy of intervention strategies organized around the different phases where they can be applied.}
    \label{fig:overview}
    \vspace{-1.5em}
\end{figure}

Indeed, LMs have been shown to introduce vulnerabilities and threats, both inadvertent and malicious, to individual users, social groups, and content integrity. 
Without social context and content control, deployed language generators have quickly derailed to racist, homophobic, hateful comments \cite{tay,racict-korean,wolf2017we,gpt-4chan}, 
compromised user privacy \cite{carlini2021extracting}, spread disinformation \cite{shao2018spread}, and even encouraged suicide \cite{gpt3}. 
Prior works have outlined these risks~\citep{factentailment, sheng-etal-2021-societal, weidinger2022taxonomy, Zhuo2023ExploringAE}, proposed taxonomies \citep{weidinger2022taxonomy}, discussed their points of origin, and advocated for future research on ethical development of LMs~\citep{10.1145/3442188.3445922,solaiman2019release}. 

However, there is little work that summarizes \textbf{actionable approaches and technical solutions} to preventing or mitigating these potential harms. 
In this survey, we present a \textbf{comprehensive, unified taxonomy} of relevant \textbf{mitigation strategies} proposed in prior literature, specifically focusing on \textbf{language generation models}. 

We organize these strategies based on where they fit in different stages of LM development: in data collection, modeling, decoding, and deployment. 
Within each of these categories, our taxonomy brings together prior works that have been treated as disjoint areas targeting different types of harms (toxic/biased language and misinformation). 
In addition
, we identify their gaps and highlight directions for future research. These include incorporating sociocultural context to produce socially-sensitive interventions, detecting and handling generations with different intents (inadvertent vs. malicious), and going beyond an English, Western/US-centric view to account for the challenges of ethics in \textit{multilingual} language generation.

\section{Background}
\label{sec:prior-surveys}
Throughout this paper, we use the term \textit{language models} (LMs) to refer to their classic definition as generative models,
which predict the next token given the preceding generated context. 
This paradigm also subsumes conditional 
LMs that depend on additional inputs via an encoder. 
We provide more details in \Aref{sec:app-back}.
\subsection{Risks in Language Generation}
Before diving into mitigation techniques (\Sref{sec:risk-mitigation}), we briefly outline potential harms that LMs can cause, following \citet{weidinger2022taxonomy}'s taxonomy.
 
\paragraph{Discrimination, Toxicity, and Exclusion: } 
The scope of linguistic diversity in human communication is enormous and is linked to personal, social, and cultural factors \citep{holmes2017introduction,gendereckert_mcconnell-ginet_2003,genderCoatesJennifer2016WMaL,socialvariation1995}. 
As such, language produced in the real world reflects sociocultural stereotypes and presuppositions that LMs can overfit to and amplify
\citep{bar2013stereotyping,zhao2017men,sun-etal-2019-mitigating},
leading to several types of harms. 
\begin{inparaenum}[(1)]
\item 
\textit{Stereotyping and discrimination} occurs when generated text reinforces discriminatory stereotypes and perpetuates biases against disadvantaged groups, based on factors like gender, race, religion, sexuality, \cite{10.1145/3442188.3445922}, and intersectional identities \citep{kimberle2017}. Evidence for this behavior has been substantially corroborated in NLP literature \citep[\textit{inter alia}]{blodgett-etal-2020-language,nadeem-etal-2021-stereoset,nozza-etal-2021-honest, liang2021towards, field-etal-2021-survey,Lin2022GenderedMH}. 
\item \textit{Toxicity} describes generated language that is offensive, threatening, violent, or otherwise harmful~\citep{gehman-etal-2020-realtoxicityprompts,DBLP:journals/corr/abs-2112-11446, 10.1145/3461702.3462624}.
It can range from overtly toxic content, such as violent hate speech, to more subtle, veiled toxicity, such as microaggressions~\citep{breitfeller-etal-2019-finding}. 
\item \textit{Exclusion} refers to the disparate performance of models across language variations. Models may fail to understand or generate ``non-standard'' dialects and sociolects, essentially excluding speakers of such variants from their user base~\citep{joshi-etal-2020-state,doi:10.1073/pnas.1915768117,winata-etal-2021-language}.
\end{inparaenum}

\paragraph{Factual Errors, Misinformation, and Disinformation: }
LMs are able to generate fluent outputs that users may easily mistake for human-written text~\citep{ippolito-etal-2020-automatic}, but such utterances 
 may be 
 factually incorrect or misleading~\citep{factentailment,openaifake,lin2021truthfulqa,11827,gpt3}. 
This can cause harm inadvertently (via misinformation) or can also be used maliciously  \citep[disinformation;][]{bradshaw2019global,Beskow2020,buchanan2021truth}. 

\paragraph{Privacy Violations:} LMs' vast training corpora often contain sensitive information, and LMs can memorize these details and generate them verbatim when prompted by users, leading to privacy violations~\citep{privacy2016kim,  Mirshghallah2020PrivacyID, Brown2022WhatDI}. LMs have been shown to leak 
personally identifiable information, such as social security numbers, phone numbers, bank account information \citep{carlini2021extracting}, and private clinical notes \citep{lehman-etal-2021-bert}; they have even leaked software code and other protected intellectual property \cite{Ippolito2022PreventingVM}. Deploying large LMs can thus pose serious security risks to people whose private information might have found its way into a model's training data. 

\noindent
\textbf{Other Underexplored
Issues: }
\citet{weidinger2022taxonomy} discuss other malicious applications, as well as the economical and environmental impacts of LMs. While extremely important, mitigating these risks requires not only technical innovation, but also the development of regulatory practices and policies in an interdisciplinary effort. We focus on algorithmic solutions in this survey, leaving this discussion for future work. 

\begin{table*}[]
\small
\begin{tabular}{p{0.15\textwidth} p{0.15\textwidth} p{0.12\textwidth} p{0.5\textwidth}}
\hline

\multirow{14}{*}{\textbf{\shortstack[l]{Application Level\\ Interventions}}} 
& \multirow{4}{*}{\shortstack[l]{Feature-based\\ Detection}} & Toxicity & Lexical features {\citep{xiang2012detecting,dadvar2012improved,burnap2015cyber,liu2015new}; n-gram features \cite{chen2012detecting, waseem2016hateful, nobata2016abusive,xu2012learning, burnap2016us}}\\
& & Misinformation & Word-Level features \citep{zhao2020reducing, king2022don} \\ \cline{2-4}
& \multirow{10}{*}{\shortstack[l]{Neural\\ Detection}} & Toxicity & Supervised: \citep{gamback2017using,pitsilis2018effective,d2020bert,xiang-etal-2021-toxccin}; Semi- and Unsupervised: \citep{Korzeniowski2019ExploitingUP, Field2020UnsupervisedDO,gnrem2021LeveragingBI} \\ 
& & Misinformation / Factuality & Supervised fake-news detection \citep{thorne-etal-2018-fever,oshikawa-etal-2020-survey,ijcai2020-0672,Zhou2020,guo-etal-2022-survey}; Factual error detection \citep{factcc, dae, frank} \\ 
& & Disinformation & Machine-generated text detection \citep{dugan-etal-2020-roft, gehrmann-etal-2019-gltr}\\ 
\hline

\multirow{12}{*}{\textbf{\shortstack[l]{Output Level\\Interventions}}} 
& \multirow{3}{*}{Reranking} &  Toxicity & Rejection sampling using toxicity detectors \citep{https://doi.org/10.48550/arxiv.2202.04173} \\ 
& & Misinformation / Factuality & Ranking using factuality classifiers \citep{Krishna2022RankGenIT,king2022don}\\ \cline{2-4}
& \multirow{9}{*}{\shortstack[l]{Controlled\\ Decoding}} & Toxicity &  Autoregressive toxic content control \citep{yang2021fudge,liu-etal-2021-dexperts, dathathri2019plug, krause-etal-2021-gedi-generative, schick2020self, lu-etal-2021-neurologic, pascual-etal-2021-plug-play, Wolf2019HuggingFacesTS}; Non-autoregressive toxic content control\citep{kumar2022gradient,mireshghallah-etal-2022-mix} \\ 
& & Privacy & Differentially private decoding \citep{Majmudar2022DifferentiallyPD}\\
& & Misinformation / Factuality &  
Autoregressive factual error control\citep{king2022don,https://doi.org/10.48550/arxiv.2112.08726}; Non-autoregressive factual error control~\citep{kumar2021controlled}  \\ \cline{2-4}
& \multirow{4}{*}{Post-processing} & Toxicity & Rewriting harmful text \citep{Pryzant_DiehlMartinez_Dass_Kurohashi_Jurafsky_Yang_2020,he-etal-2021-detect-perturb,ma-etal-2020-powertransformer} \\ 
&  & Misinformation / Factuality &  Editing factual errors~\citep{cao-etal-2020-factual,lee2022factual, factedit}  \\ 
\hline

\multirow{15}{*}{\textbf{\shortstack[l]{Model Level\\ Interventions}}} & \multirow{4}{*}{Architecture} & Misinformation / Factuality &  Attention \citep{nan-etal-2021-improving,zhu-etal-2021-enhancing}, Coreference \citep{https://doi.org/10.48550/arxiv.2109.03858};  Text Entailment \citep{DBLP:conf/acl/FalkeRUDG19,li-etal-2018-ensure}; Others \citep{wiseman-etal-2018-learning,DBLP:conf/acl/FalkeRUDG19,wan2022factpegasus}. \\ \cline{2-4}
& \multirow{11}{*}{Training} &  Toxicity &   Class-conditional
LMs \citep{keskarCTRL2019, gururangan-etal-2020-dont, chan2021cocon}; Instruction-based learning \citep{Ouyang2022TrainingLM,wei2022finetuned} \\
& & Privacy & Differential Private training \citep{kerrigan-etal-2020-differentially, Li2021LargeLM, Shi2021SelectiveDP}; Knowledge Unlearning \citep{Jang2022KnowledgeUF} \\ 
& & Misinformation / Factuality & Structured KBs \citep{wang-etal-2021-wikigraphs,10.1162/tacl_a_00476,10.1145/3512467,10.1162/tacl_a_00476,NEURIPS2020_6b493230,NEURIPS2019_f8d2e80c,izacard-grave-2021-leveraging,hossain-etal-2020-simple,NEURIPS2020_6b493230}, Retrieval-based \citep{NEURIPS2019_f8d2e80c,izacard-grave-2021-leveraging,hossain-etal-2020-simple}; Summarization \citep{huang-etal-2020-knowledge}, Translation \citep{bapna-firat-2019-non}, Dialogue models \citep{dinan2018wizard,10.1162/tacl_a_00356,zhang-etal-2020-grounded} \\ \cline{2-4}
& \multirow{6}{*}{Fine-tuning} & Discrimination \& Toxicity & Supervised fine-tuning \citep{gururangan-etal-2020-dont, chan2021cocon, https://doi.org/10.48550/arxiv.2107.13586}; RL based fine-tuning \citep{Alabdulkarim2021GoalDirectedSG, Liu2021MitigatingPB, Ouyang2022TrainingLM,Stiennon2020LearningTS}; Prompt-based learning \citep{gehman-etal-2020-realtoxicityprompts} \\
& & Exclusion & Adapting for low-resource varieties \citep{chronopoulou-etal-2020-reusing,kumar-etal-2021-machine} \\ 
\cline{2-4}
& \multirow{4}{*}{Model Editing} & Toxicity & Modifying FF layers\citep{Geva2022TransformerFL} \\ 
& & Misinformation / Factuality & Auxiliary editors to modify parameters \citep{de-cao-etal-2021-editing, mitchell2022fast}; Modify parameters associated with behavior \citep{meng2022locating, meng2023massediting} \\ 
\hline

\multirow{6}{*}{\textbf{Data}} & \multirow{3}{*}{Filtration} & Toxicity & Removing 'unwanted' words from corpus \citep{JMLR:v21:20-074, NEURIPS2020_1457c0d6, dodge2021documenting}; Removing toxic data using classifiers \citep{ngo2021mitigating} \\
& & Privacy & Filtering private/duplicate data \citep{Henderson2022PileOL, Kandpal2022DeduplicatingTD, lee-etal-2022-deduplicating}\\
\cline{2-4}
& \multirow{3}{*}{Augmentation} & Discrimination & Adding synthetically generated data \citep{dinan-etal-2020-queens,liu-etal-2020-gender,stafanovics-etal-2020-mitigating}   \\
& & Toxicity & Adding safer example data \citep{https://doi.org/10.48550/arxiv.1808.04409} \\ \hline
\end{tabular}
\caption{Strategies for mitigating various risks and harms from language models.} 
\label{tab:references}
\end{table*} 

\section{Taxonomy of Intervention Strategies}
\label{sec:risk-mitigation}

The development pipeline of a typical machine learning 
model involves several critical decisions where risks of harms can arise. Stakeholders have access to different pipeline components and therefore may employ different intervention strategies. For example, while a researcher involved in data curation can intervene before training, an application developer with limited access to a black-box model might only be able to intervene at inference. 
We present a taxonomy of intervention strategies organized by the stages of a model development life-cycle (\fref{fig:overview}), aiming to 
showcase the tools
that can be employed at different stages. We step backward through the pipeline, beginning with application-level interventions employed post-deployment and peeling back the layers through output-level interventions, model interventions, and finally ending at data-level interventions (summarized in \tref{tab:references}).

\subsection{Application Level Interventions}
\subsubsection{Harm Detection and Redaction}
\label{subsec:detection}
In order to mitigate harms at the application level, we first need to be able to \textit{detect} problematic, incorrect, and unreliable model outputs \cite{raji2020closing}. User-facing applications can employ detectors to intervene before harmful text reaches a user.
Such detectors are typically coarse, binary text classifiers, often trained for a single task, such as predicting toxicity \citep{nobata2016abusive, Davidson2017AutomatedHS,  xiang-etal-2021-toxccin}, or the factual accuracy of the outputs~\citep{factcc, dae, factqa}. 

Early approaches to building toxicity detectors focused on linear models relying on hand-designed \textit{features} based on lexicons, e.g., \href{https://hatebase.org/}{hatebase},  \citep{xiang2012detecting,dadvar2012improved,burnap2015cyber,liu2015new}, $n$-grams, capitalization/punctuation details \citep{chen2012detecting,waseem2016hateful,nobata2016abusive,xu2012learning,burnap2016us}. For misinformation detection, features like the presence of new entities or facts in generated document summaries have been employed which can indicate hallucination \citep{zhao2020reducing, king2022don}. 

Linear classifiers, while interpretable, tend to overfit to lexical features, are prone to false positives, and are easy for malicious users to bypass \citep{kurita2019towards}. Neural text classifiers, on the other hand, can incorporate contextual information and have been shown to be more robust \citep{gamback2017using, pitsilis2018effective}. 
When built by finetuning pretrained LMs instead of training from scratch, they naturally lead to even better performance 
\citep{d2020bert, xiang-etal-2021-toxccin}. Based on these models several toxicity detection tools like \href{https://perspectiveapi.com/}{Perspective API}, \href{https://beta.openai.com/docs/models/content-filter}{OpenAI content filter} or \href{https://github.com/microsoft/TOXIGEN}{ToxiGEN} are now publicly available.


To train classifiers for toxicity detection, annotated datasets in several domains have been collected for English \citep{davidson2017automated, waseem2016hateful, wiegand2018overview, pavlopoulos2017deep, mubarak2017abusive, moon2020beep}, especially to detect overtly toxic text. Human annotation efforts for more subtle toxicities like microaggressions, however, is challenging due to annotators' own biases 
~\citep{breitfeller-etal-2019-finding}. Hence, unsupervised or distantly supervised approaches have been adopted to detect them~\citep{Korzeniowski2019ExploitingUP, Field2020UnsupervisedDO,gnrem2021LeveragingBI}. Compared to English, such resources for other languages are severely lacking ~\citep{ousidhoum2019multilingual}. 

Information-related harms can arise either inadvertently (due to model errors) or deliberately (due to malicious users). 
Detecting 
manipulation
in the human-written text is an active area of research and those approaches can also be employed for machine-generated text. 
Prominent research directions include automated fact-checking, propaganda, or fake news detection for which several annotated datasets \citep{oshikawa-etal-2020-survey,ijcai2020-0672,Zhou2020,guo-etal-2022-survey, Huang2022FakingFN} and shared tasks \citep{thorne-etal-2018-fever,da-san-martino-etal-2019-findings, feldman2021proceedings} exist. 
These approaches have also been adopted to assist human fact-checkers~\citep{Shaar2021AssistingTH, Nakov2021AutomatedFF}.
However,
humans are easily fooled by machine-generated fake news
\citep{zellers2019defending,ippolito-etal-2020-automatic}. An alternate     solution is to, not find informational discrepancies, but simply detect and flag whether the text has been machine-generated \cite{gehrmann-etal-2019-gltr, dugan-etal-2020-roft, ippolito-etal-2020-automatic, mitchell2023detectgpt}, putting the onus to trust the information on the users \citep{jawahar-etal-2020-automatic}.  

To detect \textit{inadvertent} factual errors,
prior works have developed classifiers by training them to detect heuristically introduced synthetic errors in factually correct text~\cite{factcc, dae}, or question-answering errors 
using targeted QA models \cite{Scialom2021QuestEvalSA}. Being trained on synthetic data, such detectors typically do not generalize and have low human judgment correlations~\cite{frank}. 

Relying on the detectors, the most straightforward way a user-facing application can prevent harm is to not display the text at all (\textit{redacting}) or to display it with a warning sign (\textit{flagging}) \cite{xu2020recipes}. Even when the detectors are imperfect, explicitly flagging problematic outputs is still useful because it signals users to take model outputs with a grain of salt. 
However, this strategy is not always applicable: for example, in speech-based dialogue agents, ``displaying'' a warning sign is a nontrivial UX decision, and in auto-complete assistants (such as in \href{https://www.blog.google/products/gmail/subject-write-emails-faster-smart-compose-gmail/}{Gmail Smart Compose}), redacting is not an option and simply warning may not dissuade users from accepting the generated text.

\noindent \textbf{Challenges:} 
Predicting whether a text is harmful is often highly contextual and subjective. For toxicity detection, factors like region, political views, and the users' sociocultural background affect whether they perceive the text as toxic \cite{xenos2021toxicity}. Existing datasets are often biased due to their curation process \cite{dixon2018measuring, wiegand2019detection, geva2019we, sap2021annotators, factcc} and can have unreliable annotations \cite{ross2017measuring,Field2020UnsupervisedDO,frank}. Further, as with many black-box models, classifiers overfit to spurious artifacts \citep{gururangan-etal-2018-annotation,McCoy2019RightFT,kumar-etal-2019-topics} and amplify biases in their training data \citep{zhao2017men,sun-etal-2019-mitigating}. For instance, toxicity detectors have been shown to disproportionately flag African-American English (AAE) as toxic \cite{sap2019risk}. Additionally, such filters might overfit to a subset of small features, with more subtle problematic text evading such filters. \citet{Ippolito2022PreventingVM} show that blocking verbatim training data is insufficient for mitigating privacy concerns in code-generation. We discuss these issues further in \Sref{sec:future}, highlighting future directions to building finer-grained and explainable approaches for detecting harmful text.

\subsection{Output Level Interventions}
Increasingly, practitioners are building applications using LMs as APIs without explicit knowledge of how the model was trained or what training data was used.\footnote{see \url{https://gpt3demo.com/} for examples} Such APIs may vary in how much information developers can see: some allow access to all LM parameters, while black box APIs like GPT3 limit access to model outputs only. Hence, multiple solutions have been proposed for intervening at \textit{model output generation} by editing the outputs with auxiliary models or modifying decoding algorithms. 

\subsubsection{Post-Factum Editing Model Outputs}
Recent studies have explored ways to \emph{edit or revise} model-generated text to remove harmful content. Text editing is a decades-old subfield of NLP that has traditionally focused on fixing errors in machine translation~\citep{chollampatt2020can, Simard2007RuleBasedTW, chatterjee-etal-2020-findings} or grammar in human-written text~\citep{Wang2020ACS}. While many approaches in this area are applicable to post-editing LM outputs, in this survey, we highlight recent work related to \textit{rewriting harmful text}. 

The first set of works treats the task of rewriting as a sequence labeling task, where each token in the output sequence is either substituted, deleted, or kept the same~\citep{Pryzant_DiehlMartinez_Dass_Kurohashi_Jurafsky_Yang_2020,he-etal-2021-detect-perturb}. This, however, can be limiting when the entire output needs rewriting.
For text-to-text tasks, like translation, summarization, etc. which are trained with parallel data, the same data can be adapted to train
an editing model 
by converting source-target pairs to source-\textit{output}-target triplets using model-generated \textit{outputs} for each source, along with an additional signal indicating errors (obtained using automatic evaluators or human judgment). 
For more open-ended tasks, prior works explored unsupervised solutions for bias correction~\citep{ma-etal-2020-powertransformer} and semi-supervised methods to correct factual errors~\citep{cao-etal-2020-factual,lee2022factual, factedit}. Such methods create synthetic data by inducing errors in clean text and train a model to correct them.

\subsubsection{Decoding Methods}
\label{subsec:decoding}
Several search and sampling algorithms have been introduced recently to improve the quality of LM-generated text~\citep{graves2012sequence,fan-etal-2018-hierarchical,Holtzman2020The,typical}. In parallel, works on controlling decoding algorithms to promote or demote specific properties in the output text have been developed~\citep{CTGsurvey}. 

The decoding controls are auxiliary models measuring if the generated text is harmful implemented similarly to the detectors 
we discussed in \Sref{subsec:detection}, such as toxicity/bias classifiers~\citep{dathathri2019plug,krause-etal-2021-gedi-generative,liu-etal-2021-dexperts}, factuality metrics~\citep{factcc, dae}.
A simple way to use the detectors is \textit{rejection sampling} or \textit{reranking}: for a given input, multiple outputs are generated and then reranked using detector scores to discard dubious outputs~\citep{Krishna2022RankGenIT, king2022don}. However, this is often intractable for complex phenomena like factual accuracy of a text or when using multiple controls, since all the generated candidates might be rejected.

To tackle these issues, a class of algorithms that we call \textit{guided-autoregressive decoding} aims to incorporate control by modifying output distributions at every decoding step. One branch of work adopts \textit{logical} controls, where developers directly specify sets of words that should (or not) appear in the output~\citep{lu-etal-2021-neurologic,pascual-etal-2021-plug-play}. \citet{Wolf2019HuggingFacesTS} apply this method to zero out the probabilities of offensive terms,  ~\citet{king2022don,https://doi.org/10.48550/arxiv.2112.08726} improve factual accuracy of generated text by up-weighting generation probabilities of entities present in the source, and \citet{Majmudar2022DifferentiallyPD} apply it for differentially private decoding. A second branch of work composes the LM likelihood with the probabilities from the detectors, to up-weight or down-weight the token probabilities at each decoding step~\citep{yang2021fudge,liu-etal-2021-dexperts,dathathri2019plug,krause-etal-2021-gedi-generative,schick2020self}.

More recent work has also explored ways to induce sentence-level control via \textit{non-autoregressive controlled decoding}. These algorithms incorporate control using Monte Carlo Markov Chain (MCMC) techniques~\citep{hoang-etal-2017-towards,qin-etal-2020-back,mireshghallah-etal-2022-mix}, in which a full sequence is initialized and iteratively updated. They have been applied for reducing toxicity~\citep{kumar2022gradient}, and improving fidelity in translation systems~\citep{kumar2021controlled}. While promising, these techniques suffer from slower decoding speed 
and need further exploration to be practically used. 

\paragraph{Challenges}
Decoding interventions rely on accurate detectors, hence challenges in designing robust detectors (\Sref{subsec:detection}) also impact decoding algorithms. For example, ~\citet{xu-etal-2021-detoxifying} show that toxicity avoidance algorithms refrain from generating AAE, thereby causing another harm (exclusion) while trying to address the first (toxicity). Also, detecting misinformation and factuality can be extremely hard using simple detectors that do not provide a useful signal to guide the decoding process, so prior works have primarily employed heuristics. 
Finally, controlled decoding algorithms are double-edged in that controls can be reversed by malicious users to inflict harm---to generate hateful messages, or to do targeted manipulation by copying users' personas. However, this risk should not discourage research in decoding algorithms; rather, research on detecting such malicious uses should be conducted in parallel.

\subsection{Model Level Interventions}
Several recent studies have provided evidence that certain optimization procedures can result in harmful generations downstream~\citep{Hall2022ASS,taori2022data}.
In this section, we describe approaches that modify LM parameters to prevent such generations by either architecture/training interventions or finetuning/model editing interventions.

\subsubsection{Architecture and Training Algorithms}
Closely related to applying control at inference time are class-conditioned LMs, which are trained to depend on "control codes" via an additional input ~\citep{keskarCTRL2019,gururangan-etal-2020-dont,chan2021cocon}. When trained with data annotated for toxicity or bias, these LMs can be prompted to avoid those outputs. Another recently popularized paradigm in LM training is \textit{instruction-based learning}, where in addition to the objective to predict the next token, models are also trained to solve NLP tasks with instructions written in natural language~\citep{wei2022finetuned,sanh2022multitask}. Providing explicit instructions to not generate harmful text has shown some promise~\citep{Ouyang2022TrainingLM,wei2022finetuned} and is an interesting avenue for future work.

In text-to-text tasks like summarization, the goal is to produce text that is factually consistent with the input without hallucinating information. An LM, however, is typically not constrained to predict tokens grounded in verifiable knowledge, which can lead to misinformation. 
Thus, several studies explore modifying LM training
objectives to \textit{incorporate factual information} using either knowledge bases (KBs) or graphs~\citep{10.1145/3512467}: each token prediction is scored not only on its likelihood given context, but also on whether the generation is grounded in facts in the KBs~\citep{wang-etal-2021-wikigraphs}.\footnote{Knowledge-augmented LMs is a rich field where most existing work focuses on masked LMs~\citep{zhu-etal-2022-knowledge} for solving understanding tasks. Here we highlight papers on generation.} 

However, existing KBs are limited in size as manually curating them is an arduous and expensive process.
As an alternative, \citet{10.1162/tacl_a_00476} propose using automatically generated KBs to train LMs. In contrast, \citet{NEURIPS2020_6b493230,NEURIPS2019_f8d2e80c,izacard-grave-2021-leveraging} use unstructured text as knowledge. Known as \textit{retrieval-augmented LMs}, they are trained with a two-stage approach of first retrieving a document from an unstructured source like Wikipedia and using it as additional context for generation, essentially providing evidence for the LM-generated text. \citet{wang-etal-2021-retrieval-enhanced,ji-etal-2020-language} follow a similar approach to embed commonsense knowledge in LMs.
These existing solutions have been used to tackle content-related harms like factual consistency in generated text \cite{huang-etal-2020-knowledge, bapna-firat-2019-non, dinan2018wizard,10.1162/tacl_a_00356} but future work in reducing discrimination and toxicity in LMs may also benefit from KBs that encode social~\citep{chang-etal-2020-incorporating}, cultural,~\citep{hershcovich-etal-2022-challenges}, and moral norms~\citep{hendrycks2021ethics,Jiang2021DelphiTM}. Such LMs augmented with external knowledge can also be dynamically updated by modifying the knowledge source at test time with new information~\citep{Khandelwal2020Generalization,he-etal-2021-efficient}. 

While external knowledge helps provides context, models may not rely on them and still hallucinate. 
To explicitly control for context, recent studies have explored (1) modifying attention mechanisms to specifically capture relationships between entities~\citep{nan-etal-2021-improving,zhu-etal-2021-enhancing}, (2) improving coreference to mitigate gender bias in translation~\citep{https://doi.org/10.48550/arxiv.2109.03858}, and (3) using text entailment to develop loss functions to improve fidelity~\citep{DBLP:conf/acl/FalkeRUDG19,li-etal-2018-ensure}. Some other notable directions in this space involve fact-aware pretraining~\citep{DBLP:conf/acl/FalkeRUDG19,wan2022factpegasus} and structured learning frameworks~\citep{wiseman-etal-2018-learning}.    


Finally, to reduce privacy risks in LMs that memorize user information without sacrificing model capabilities, 
most prominent solutions are based on differentially private (DP) learning~\citep{kerrigan-etal-2020-differentially,Shi2021SelectiveDP}.
DP can provide provable guarantees on the privacy-utility trade-off,
however, it requires the LMs to be retrained for each private information that needs to be removed and be quite expensive. 

\subsubsection{Finetuning and Model Editing}
Designing and training models from scratch to mitigate harms can incur heavy
environmental and 
resource costs. In contrast, an alternative branch of work has developed methods for \textit{modifying the model parameters} of already-trained LMs, which requires much fewer resources. 
An elementary way of doing this is \textit{finetuning} (a subset of) an LM's parameters on small, curated datasets that contain a well-balanced proportion of data for various demographics and filtered for nontoxicity~\citep{gururangan-etal-2020-dont,chan2021cocon,https://doi.org/10.48550/arxiv.2107.13586}. Such balanced and filtered data encourage models correct biases learned from skewed and toxic training data, resulting in safer generated text.

\textit{Prompt-tuning} based methods~\citep{https://doi.org/10.48550/arxiv.2202.04173} have also shown some success where instead of fine-tuning all the parameters, a prompt (using a small set of parameters) is learned without modifying the rest of the model to perform a task. This paradigm uses the generative power of large LMs, while simultaneously nudging the distribution of generated text toward less harmful content. 
These approaches have successfully been used to reduce toxicity~\citep{gehman-etal-2020-realtoxicityprompts} and exclusion~\citep{chronopoulou-etal-2020-reusing,kumar-etal-2021-machine}. However, finetuning or prompt-tuning on a small dataset may lead to overfitting reducing the general purpose utility of LMs.


Finetuning LMs with \textit{reinforcement learning} (RL) has been suggested as a better alternative~\citep{Alabdulkarim2021GoalDirectedSG,Liu2021MitigatingPB,Ouyang2022TrainingLM,Stiennon2020LearningTS,https://doi.org/10.48550/arxiv.2112.08726,https://doi.org/10.48550/arxiv.2210.01241} for training modern LMs. RL models do not require carefully balanced datasets and can instead learn from discrete rewards such as human feedback~\citep{Sun2020LAMOLLM,Ouyang2022TrainingLM} or auxiliary model-based feedback \citep{Perez2022RedTL}. It has been shown to reduce toxic text generated by the models~\cite{Bai2022TrainingAH} and to encourage models to generate more factual text \citep{Mao2020ConstrainedAS, Stiennon2020LearningTS}.

Another less-explored but more computationally practical alternative to finetuning is \textit{model surgery or editing}, which 
identifies a specific set of neurons that contribute to harmful generations. Culling such parameters has been shown to reduce toxicity \citep{Geva2022TransformerFL}. In a similar vein, \citet{de-cao-etal-2021-editing,mitchell2022fast, meng2022locating, meng2023massediting} systematically edit model parameters to revise facts memorized by the model. \citet{de-cao-etal-2021-editing,mitchell2022fast} use auxiliary editor networks to predict updates to model parameters constrained to revise a fact without changing other facts. Alternatively, \citet{meng2022locating, meng2023massediting} use interpretability techniques to identify parameters associated with memorizing said facts and edit them locally to revise them. 

\paragraph{Challenges}
The biggest argument against mitigation techniques involving training LMs from scratch or augmenting them with knowledge is its cost, making these interventions infeasible for most researchers and practitioners. However, even for organizations with access to large computing resources, research on training safer LMs lags behind research on training ever-larger LMs on raw data. We attribute this to the difficulty of curating KBs, as well as the decreased training and inference speed that comes with such modifications. Finetuning, on the other hand, is less costly but may reduce the general utility of the LMs and has not been shown to be useful in reducing information-related harms. 
Future work may benefit from drawing on continual ~\citep{dhingra-etal-2022-time} and reinforcement learning~\citep{Ouyang2022TrainingLM} techniques for more practical solutions for large models. 

\subsection{Data Level Interventions}
Training any machine learning model requires data, so a natural approach to creating fairer, more reliable LMs is carefully creating balanced training sets that are broadly representative of different worldviews. This requires dedicated and expensive efforts in data curation~\citep{10.1145/3442188.3445918,10.1145/3351095.3372829,Kammoun_2022} and novel data pipelines~\citep{denton2020bringing}.
Existing works tackling this issue 
devise semi-automated solutions, which we categorize as follows.

\subsubsection{Data Filtration}
This simple technique involves removing problematic documents from the training corpus.
As training sets can be extremely large, sophisticated neural filters can be prohibitively slow to apply. Hence, most work has utilized simple filters, such as the presence of "unwanted" words~\citep{JMLR:v21:20-074} or the predictions of linear classifiers~\citep{NEURIPS2020_1457c0d6}. To mitigate privacy violations, \citet{Henderson2022PileOL} construct clean training data by filtering private information and \citet{Kandpal2022DeduplicatingTD, lee-etal-2022-deduplicating} filter duplicate training data. 

Due to their simplistic setup, these approaches admit many false negatives (failing to detect documents with subtle toxicity) and false positives (erroneously flagging documents that discuss sensitive topics and use hateful speech as examples; additionally, removing data from different dialects like AAE), unintentionally exacerbating risks of marginalization and exclusion
~\citep{dodge2021documenting}). 
Alternatively, \citet{ngo2021mitigating} train an LM on raw data, then feed the LM manually-curated toxic prompts and filter out documents to which the LM assigns high probability, and then retrain the LM on the filtered corpus.

\subsubsection{Data Augmentation}
While data filtration aims to remove problematic training samples, data augmentation aims to offset the effect of problematic data by \textit{adding} safer/healthier examples to existing datasets. \citet{https://doi.org/10.48550/arxiv.1808.04409} explore adding counterspeech (comments that counter
the hateful or harmful speech) to datasets in order to balance out the hate speech already present in web data. Augmentation with synthetically generated data has also been explored for gender bias mitigation in dialogue~\citep{dinan-etal-2020-queens,liu-etal-2020-gender} and translation models~\citep{stafanovics-etal-2020-mitigating}. 

\paragraph{Challenges}
Since language, identity, and society are tightly intertwined, aggressive data filtering methods risk further imbalancing already imbalanced data. Besides, models trained on filtered data may still degrade when toxic inputs are provided to it. Further, while data augmentation methods have merit, these methods are extremely difficult to large scale. Finally, data interventions are primarily designed to address population-centric risks such as discrimination, toxicity, and, to an extent, exclusion and privacy---but not factuality which is a by-product of training. 
It is challenging to define~\citep{fever-2022-fact} and detect unsupported facts~\citep{ANSAR2021100052} in the wild, making data interventions insufficient for addressing misinformation and factuality-related harms.

\section{Discussion and Open Challenges}
\label{sec:future}
Though the interventions strategies we discuss achieve some success, many risks of LMs are still not well understood.
Below we discuss open problems and avenues for future work to encourage the development of safer LMs.

\paragraph{Where should one intervene?}
Different stakeholders are involved in different model development phases with varying access to resources. As a result, intervention strategies are different depending on the stakeholder. A significant chunk of the responsibility to develop safer LMs falls on researchers and organizations with access to substantial resources who can implement data or modeling interventions. In contrast, practitioners building applications on top of LMs may have access to neither the training data nor the computational resources required to design and train safe LMs. In such cases, flagging and decoding approaches are more practical. 
In practice, a combination of multiple interventions may be required to both cover a wide array of risks and improve robustness.

\paragraph{Evolving risks in the ChatGPT era:} LMs are seeing tremendous, rapid growth; larger models are being released every few months 
\citep{Shoeybi2019MegatronLMTM, NEURIPS2020_1457c0d6,https://doi.org/10.48550/arxiv.2205.01068, Zeng2022GLM130BAO} and deployed in user-facing applications. 
Many recent LMs like OpenAI's ChatGPT have garnered attention beyond the research community, impacting a range of fields and crossing geographical and language barriers to reach users all over the world \cite{chatgpt-india, chatgpt-india2, chatgpt-korea}. In such a fast-moving ecosystem, it is ever more essential to proactively study and mitigate LMs' potential harms. Risk mitigation research tends to lag behind model development and is often considered as an afterthought. Though behaviors may emerge unpredictably \cite{emergent_abilities}, as we outline in this survey, intervention strategies can and should be applied at different stages of model development to reduce the potential for these influential LMs to cause harm. 

\paragraph{Risks exist in LMs in all languages:}
Most research on large LMs, their uses, and their risks is Western-centric and primarily conducted on the English language. 
However, while a few studies have been conducted on detecting harmful text in non-English datasets~\citep[\textit{inter alia}]{ousidhoum-etal-2019-multilingual,leite-etal-2020-toxic,burtenshaw-kestemont-2021-dutch,9435584,costa2022toxicity}, research on mitigation in non-English settings is lagging~\citep{pamungkas2021towards}. Further, 
the definitions of risks themselves change with different context and across cultures.
Hence, there is a dire need to develop cross-cultural, cross-lingual analyses as well as mitigation tools.

\paragraph{Harm detection beyond simple classifiers} 
Many of the shortcomings of interventions are at their root due to poorly defined risk detection methods. Current detection methods are primarily binary classifiers on various axes like toxicity and factuality, but we recommend researchers and practitioners to move beyond simplistic coarse classifiers and towards more fine-grained \citep{xiang-etal-2021-toxccin, dae, da-san-martino-etal-2020-semeval}, interpretable \citep{pmlr-v70-koh17a, han-tsvetkov-2020-fortifying,han-tsvetkov-2021-influence-tuning}, and explainable \citep{frank, gehrmann-etal-2019-gltr} harm detectors to support better harm mitigation strategies
\citep{lipton2018mythos,Jacovi2021FormalizingTI}. 


\paragraph{Systematic evaluation frameworks for mitigation strategies}
Though LM performance is usually systematically evaluated through benchmarks~\citep{wang2018glue,superglue,bigbench}, practices for evaluating harms in LM-generated text or the effectiveness of mitigation strategies are not. 
While there is an emerging body of work dedicated to benchmarking LM harms~\citep{rauh2022characteristics}, the space of potential harms is huge and intersectional, and existing work only covers a fraction of it.
Developing a suite of evaluations or augmenting existing generation benchmarks~\citep{mille2021automatic} with axes of risk evaluations~\citep{ribeiro-etal-2020-beyond} will encourage the development of holistic solutions, bridging discrimination/toxicity and information-related harms---two related directions in which researchers have often developed similar solutions. 

\section{Conclusion}
We present a survey of practical methods and techniques for addressing the societal harms and safety risks of language generation models. Our structured taxonomy covers a wide variety of interventions at different stages of the model development pipeline to mitigate harms. This work bridges multiple strands of research and presents an actionable overview on methods for preventing harms from language generation models.

\section*{Limitations}
The goal of this survey was to present current research on  analyzing and mitigating harms of language generation. There are multiple documented and anticipated harms that these models perpetuate, and it is not feasible to address intervention strategies for each of them. We aimed to generalize multiple proposed solutions and present them in a structured form, considering a few popularly studied harms as case studies. Inevitably, certain harms and their mitigation strategies might not have been considered for this survey.

Current research in this field is nascent but fast-moving. While this survey enlists techniques and approaches that are popular now, there is a potential for them to be replaced with newer research. We anticipate that this survey may need to be updated or even redone to incorporate new research.


\section*{Ethics Statement}
In this survey, we present and discuss various risk analyses and intervention strategies to prevent societal harms from LMs. We also comment on common themes across approaches for detecting and resolving population-centric harms (such as toxicity and discrimination) and misinformation-related harms, and we recommend future work combining them. First, many datasets and resources we discuss may contain biases, and using them in downstream applications can lead to risks as we have outlined. Second, many techniques we discuss have limitations or are known to exacerbate other kinds of harms~\citep{xia-etal-2020-demoting}, and thus, applying them to newer problems may lead to unseen issues. Finally, the interventions we identify to raise general awareness have the potential for misuse: a malicious user can further imbalance the data to train even \href{https://huggingface.co/ykilcher/gpt-4chan}{more harmful models}, use the models and decoding algorithms to generate fake news, and target marginalized populations. This, however, should not discourage the development of mitigation strategies; rather, more work should be done to detect and ban malicious users. This requires not only technological solutions in NLP, but also in social science, social network analysis, and public policy. 

\section*{Acknowledgements}
We gratefully acknowledge support from NSF CAREER Grant No.~IIS2142739, NSF grants No.~IIS2125201, IIS2040926, IIS2203097, Workhuman, and an Alfred P. Sloan Foundation Fellowship. 
S.K. gratefully acknowledges support from a Google PhD Fellowship.

\bibliography{anthology,custom}

\begin{thebibliography}{254}
\expandafter\ifx\csname natexlab\endcsname\relax\def\natexlab#1{#1}\fi

\bibitem[{Abid et~al.(2021)Abid, Farooqi, and Zou}]{10.1145/3461702.3462624}
Abubakar Abid, Maheen Farooqi, and James Zou. 2021.
\newblock \href {https://doi.org/10.1145/3461702.3462624} {\emph{Persistent
  Anti-Muslim Bias in Large Language Models}}, page 298–306. Association for
  Computing Machinery, New York, NY, USA.

\bibitem[{Adiwardana et~al.(2020)Adiwardana, Luong, So, Hall, Fiedel,
  Thoppilan, Yang, Kulshreshtha, Nemade, Lu et~al.}]{adiwardana2020towards}
Daniel Adiwardana, Minh-Thang Luong, David~R So, Jamie Hall, Noah Fiedel, Romal
  Thoppilan, Zi~Yang, Apoorv Kulshreshtha, Gaurav Nemade, Yifeng Lu, et~al.
  2020.
\newblock Towards a human-like open-domain chatbot.
\newblock \emph{arXiv preprint arXiv:2001.09977}.

\bibitem[{Aghajanyan et~al.(2022)Aghajanyan, Okhonko, Lewis, Joshi, Xu, Ghosh,
  and Zettlemoyer}]{aghajanyan2022htlm}
Armen Aghajanyan, Dmytro Okhonko, Mike Lewis, Mandar Joshi, Hu~Xu, Gargi Ghosh,
  and Luke Zettlemoyer. 2022.
\newblock \href {https://openreview.net/forum?id=P-pPW1nxf1r} {{HTLM}:
  Hyper-text pre-training and prompting of language models}.
\newblock In \emph{International Conference on Learning Representations}.

\bibitem[{Alabdulkarim et~al.(2021)Alabdulkarim, Li, Martin, and
  Riedl}]{Alabdulkarim2021GoalDirectedSG}
Amal Alabdulkarim, Winston Wai-Tai Li, Lara~J. Martin, and Mark~O. Riedl. 2021.
\newblock Goal-directed story generation: Augmenting generative language models
  with reinforcement learning.
\newblock \emph{ArXiv}, abs/2112.08593.

\bibitem[{Aly et~al.(2022)Aly, Christodoulopoulos, Cocarascu, Guo, Mittal,
  Schlichtkrull, Thorne, and Vlachos}]{fever-2022-fact}
Rami Aly, Christos Christodoulopoulos, Oana Cocarascu, Zhijiang Guo, Arpit
  Mittal, Michael Schlichtkrull, James Thorne, and Andreas Vlachos, editors.
  2022.
\newblock \href {https://aclanthology.org/2022.fever-1.0} {\emph{Proceedings of
  the Fifth Fact Extraction and VERification Workshop (FEVER)}}. Association
  for Computational Linguistics, Dublin, Ireland.

\bibitem[{Ansar and Goswami(2021)}]{ANSAR2021100052}
Wazib Ansar and Saptarsi Goswami. 2021.
\newblock \href {https://doi.org/https://doi.org/10.1016/j.jjimei.2021.100052}
  {Combating the menace: A survey on characterization and detection of fake
  news from a data science perspective}.
\newblock \emph{International Journal of Information Management Data Insights},
  1(2):100052.

\bibitem[{Bai et~al.(2022)Bai, Jones, Ndousse, Askell, Chen, DasSarma, Drain,
  Fort, Ganguli, Henighan, Joseph, Kadavath, Kernion, Conerly, El-Showk,
  Elhage, Hatfield-Dodds, Hernandez, Hume, Johnston, Kravec, Lovitt, Nanda,
  Olsson, Amodei, Brown, Clark, McCandlish, Olah, Mann, and
  Kaplan}]{Bai2022TrainingAH}
Yuntao Bai, Andy Jones, Kamal Ndousse, Amanda Askell, Anna Chen, Nova DasSarma,
  Dawn Drain, Stanislav Fort, Deep Ganguli, T.~J. Henighan, Nicholas Joseph,
  Saurav Kadavath, John Kernion, Tom Conerly, Sheer El-Showk, Nelson Elhage,
  Zac Hatfield-Dodds, Danny Hernandez, Tristan Hume, Scott Johnston, Shauna
  Kravec, Liane Lovitt, Neel Nanda, Catherine Olsson, Dario Amodei, Tom~B.
  Brown, Jack Clark, Sam McCandlish, Christopher Olah, Benjamin Mann, and Jared
  Kaplan. 2022.
\newblock Training a helpful and harmless assistant with reinforcement learning
  from human feedback.
\newblock \emph{ArXiv}, abs/2204.05862.

\bibitem[{Balachandran et~al.(2022)Balachandran, Hajishirzi, Cohen, and
  Tsvetkov}]{factedit}
Vidhisha Balachandran, Hannaneh Hajishirzi, William Cohen, and Yulia Tsvetkov.
  2022.
\newblock Correcting diverse factual errors in abstractive summarization via
  post-editing and language model infilling.
\newblock In \emph{Proceedings of the 2022 Conference on Empirical Methods in
  Natural Language Processing (EMNLP)}.

\bibitem[{Bapna and Firat(2019)}]{bapna-firat-2019-non}
Ankur Bapna and Orhan Firat. 2019.
\newblock \href {https://doi.org/10.18653/v1/N19-1191} {Non-parametric
  adaptation for neural machine translation}.
\newblock In \emph{Proceedings of the 2019 Conference of the North {A}merican
  Chapter of the Association for Computational Linguistics: Human Language
  Technologies, Volume 1 (Long and Short Papers)}, pages 1921--1931,
  Minneapolis, Minnesota. Association for Computational Linguistics.

\bibitem[{Bar-Tal et~al.(2013)Bar-Tal, Graumann, Kruglanski, and
  Stroebe}]{bar2013stereotyping}
Daniel Bar-Tal, Carl~F Graumann, Arie~W Kruglanski, and Wolfgang Stroebe. 2013.
\newblock \emph{Stereotyping and prejudice: Changing conceptions}.
\newblock Springer Science \& Business Media.

\bibitem[{Bender et~al.(2021)Bender, Gebru, McMillan-Major, and
  Shmitchell}]{10.1145/3442188.3445922}
Emily~M. Bender, Timnit Gebru, Angelina McMillan-Major, and Shmargaret
  Shmitchell. 2021.
\newblock \href {https://doi.org/10.1145/3442188.3445922} {On the dangers of
  stochastic parrots: Can language models be too big?}
\newblock In \emph{Proceedings of the 2021 ACM Conference on Fairness,
  Accountability, and Transparency}, FAccT '21, page 610–623, New York, NY,
  USA. Association for Computing Machinery.

\bibitem[{Beskow(2020)}]{Beskow2020}
David Beskow. 2020.
\newblock \href {https://doi.org/10.1184/R1/12303089.v1} {{Finding and
  Characterizing Information Warfare Campaigns}}.

\bibitem[{Bickmore et~al.(2018)Bickmore, Trinh, Olafsson, O'Leary, Asadi,
  Rickles, and Cruz}]{11827}
Timothy Bickmore, Ha~Trinh, Stefan Olafsson, Teresa O'Leary, Reza Asadi,
  Nathaniel Rickles, and Ricardo Cruz. 2018.
\newblock \href {https://doi.org/10.2196/11510} {Patient and consumer safety
  risks when using conversational assistants for medical information: an
  observational study of siri, alexa, and google assistant}.
\newblock \emph{J Med Internet Res}, 20:e11510.

\bibitem[{{BIG-bench collaboration}(2022)}]{bigbench}
{BIG-bench collaboration}. 2022.
\newblock Beyond the imitation game: Measuring and extrapolating the
  capabilities of language models.

\bibitem[{Blodgett et~al.(2020)Blodgett, Barocas, Daum{\'e}~III, and
  Wallach}]{blodgett-etal-2020-language}
Su~Lin Blodgett, Solon Barocas, Hal Daum{\'e}~III, and Hanna Wallach. 2020.
\newblock \href {https://doi.org/10.18653/v1/2020.acl-main.485} {Language
  (technology) is power: A critical survey of {``}bias{''} in {NLP}}.
\newblock In \emph{Proceedings of the 58th Annual Meeting of the Association
  for Computational Linguistics}, pages 5454--5476, Online. Association for
  Computational Linguistics.

\bibitem[{Bogoradnikova et~al.(2021)Bogoradnikova, Makhnytkina, Matveev,
  Zakharova, and Akulov}]{9435584}
Darya Bogoradnikova, Olesia Makhnytkina, Anton Matveev, Anastasia Zakharova,
  and Artem Akulov. 2021.
\newblock \href {https://doi.org/10.23919/FRUCT52173.2021.9435584}
  {Multilingual sentiment analysis and toxicity detection for text messages in
  russian}.
\newblock In \emph{2021 29th Conference of Open Innovations Association
  (FRUCT)}, pages 55--64.

\bibitem[{Bradshaw and Howard(2019)}]{bradshaw2019global}
Samantha Bradshaw and Philip~N Howard. 2019.
\newblock The global disinformation order: 2019 global inventory of organised
  social media manipulation.

\bibitem[{Breitfeller et~al.(2019)Breitfeller, Ahn, Jurgens, and
  Tsvetkov}]{breitfeller-etal-2019-finding}
Luke Breitfeller, Emily Ahn, David Jurgens, and Yulia Tsvetkov. 2019.
\newblock \href {https://doi.org/10.18653/v1/D19-1176} {Finding
  microaggressions in the wild: A case for locating elusive phenomena in social
  media posts}.
\newblock In \emph{Proceedings of the 2019 Conference on Empirical Methods in
  Natural Language Processing and the 9th International Joint Conference on
  Natural Language Processing (EMNLP-IJCNLP)}, pages 1664--1674, Hong Kong,
  China. Association for Computational Linguistics.

\bibitem[{Brown et~al.(2022)Brown, Lee, Mireshghallah, Shokri, and
  Tram{\`e}r}]{Brown2022WhatDI}
Hannah Brown, Katherine Lee, FatemehSadat Mireshghallah, R.~Shokri, and Florian
  Tram{\`e}r. 2022.
\newblock What does it mean for a language model to preserve privacy?
\newblock \emph{2022 ACM Conference on Fairness, Accountability, and
  Transparency}.

\bibitem[{Brown et~al.(2020)Brown, Mann, Ryder, Subbiah, Kaplan, Dhariwal,
  Neelakantan, Shyam, Sastry, Askell, Agarwal, Herbert-Voss, Krueger, Henighan,
  Child, Ramesh, Ziegler, Wu, Winter, Hesse, Chen, Sigler, Litwin, Gray, Chess,
  Clark, Berner, McCandlish, Radford, Sutskever, and
  Amodei}]{NEURIPS2020_1457c0d6}
Tom Brown, Benjamin Mann, Nick Ryder, Melanie Subbiah, Jared~D Kaplan, Prafulla
  Dhariwal, Arvind Neelakantan, Pranav Shyam, Girish Sastry, Amanda Askell,
  Sandhini Agarwal, Ariel Herbert-Voss, Gretchen Krueger, Tom Henighan, Rewon
  Child, Aditya Ramesh, Daniel Ziegler, Jeffrey Wu, Clemens Winter, Chris
  Hesse, Mark Chen, Eric Sigler, Mateusz Litwin, Scott Gray, Benjamin Chess,
  Jack Clark, Christopher Berner, Sam McCandlish, Alec Radford, Ilya Sutskever,
  and Dario Amodei. 2020.
\newblock \href
  {https://proceedings.neurips.cc/paper/2020/file/1457c0d6bfcb4967418bfb8ac142f64a-Paper.pdf}
  {Language models are few-shot learners}.
\newblock In \emph{Advances in Neural Information Processing Systems},
  volume~33, pages 1877--1901. Curran Associates, Inc.

\bibitem[{Buchanan et~al.(2021)Buchanan, Lohn, Musser, and
  Sedova}]{buchanan2021truth}
Ben Buchanan, Andrew Lohn, Micah Musser, and Katerina Sedova. 2021.
\newblock Truth, lies, and automation.

\bibitem[{Burnap and Williams(2015)}]{burnap2015cyber}
Pete Burnap and Matthew~L Williams. 2015.
\newblock Cyber hate speech on twitter: An application of machine
  classification and statistical modeling for policy and decision making.
\newblock \emph{Policy \& internet}, 7(2):223--242.

\bibitem[{Burnap and Williams(2016)}]{burnap2016us}
Pete Burnap and Matthew~L Williams. 2016.
\newblock Us and them: identifying cyber hate on twitter across multiple
  protected characteristics.
\newblock \emph{EPJ Data science}, 5:1--15.

\bibitem[{Burtenshaw and Kestemont(2021)}]{burtenshaw-kestemont-2021-dutch}
Ben Burtenshaw and Mike Kestemont. 2021.
\newblock \href {https://aclanthology.org/2021.bucc-1.10} {A {D}utch dataset
  for cross-lingual multilabel toxicity detection}.
\newblock In \emph{Proceedings of the 14th Workshop on Building and Using
  Comparable Corpora (BUCC 2021)}, pages 75--79, Online (Virtual Mode). INCOMA
  Ltd.

\bibitem[{Cao et~al.(2020)Cao, Dong, Wu, and Cheung}]{cao-etal-2020-factual}
Meng Cao, Yue Dong, Jiapeng Wu, and Jackie Chi~Kit Cheung. 2020.
\newblock \href {https://doi.org/10.18653/v1/2020.emnlp-main.506} {Factual
  error correction for abstractive summarization models}.
\newblock In \emph{Proceedings of the 2020 Conference on Empirical Methods in
  Natural Language Processing (EMNLP)}, pages 6251--6258, Online. Association
  for Computational Linguistics.

\bibitem[{Carlini et~al.(2021)Carlini, Tramer, Wallace, Jagielski,
  Herbert-Voss, Lee, Roberts, Brown, Song, Erlingsson
  et~al.}]{carlini2021extracting}
Nicholas Carlini, Florian Tramer, Eric Wallace, Matthew Jagielski, Ariel
  Herbert-Voss, Katherine Lee, Adam Roberts, Tom Brown, Dawn Song, Ulfar
  Erlingsson, et~al. 2021.
\newblock Extracting training data from large language models.
\newblock In \emph{30th USENIX Security Symposium (USENIX Security 21)}, pages
  2633--2650.

\bibitem[{Chambers(1995)}]{socialvariation1995}
J.~K. Chambers. 1995.
\newblock \emph{Sociolinguistic theory: linguistic variation and its social
  significance}.
\newblock Oxford.

\bibitem[{Chan et~al.(2021)Chan, Ong, Pung, Zhang, and Fu}]{chan2021cocon}
Alvin Chan, Yew-Soon Ong, Bill Pung, Aston Zhang, and Jie Fu. 2021.
\newblock Cocon: A self-supervised approach for controlled text generation.
\newblock In \emph{Proc. ICLR}.

\bibitem[{Chang et~al.(2020)Chang, Liu, Gopalakrishnan, Hedayatnia, Zhou, and
  Hakkani-Tur}]{chang-etal-2020-incorporating}
Ting-Yun Chang, Yang Liu, Karthik Gopalakrishnan, Behnam Hedayatnia, Pei Zhou,
  and Dilek Hakkani-Tur. 2020.
\newblock \href {https://doi.org/10.18653/v1/2020.deelio-1.9} {Incorporating
  commonsense knowledge graph in pretrained models for social commonsense
  tasks}.
\newblock In \emph{Proceedings of Deep Learning Inside Out (DeeLIO): The First
  Workshop on Knowledge Extraction and Integration for Deep Learning
  Architectures}, pages 74--79, Online. Association for Computational
  Linguistics.

\bibitem[{ChatGPT and Perlman(2022)}]{ChatGPT2022TheIO}
Open AI's~Assistant ChatGPT and Andrew~M. Perlman. 2022.
\newblock The implications of openai’s assistant for legal services and
  society.
\newblock \emph{SSRN Electronic Journal}.

\bibitem[{Chatterjee et~al.(2020)Chatterjee, Freitag, Negri, and
  Turchi}]{chatterjee-etal-2020-findings}
Rajen Chatterjee, Markus Freitag, Matteo Negri, and Marco Turchi. 2020.
\newblock \href {https://aclanthology.org/2020.wmt-1.75} {Findings of the {WMT}
  2020 shared task on automatic post-editing}.
\newblock In \emph{Proceedings of the Fifth Conference on Machine Translation},
  pages 646--659, Online. Association for Computational Linguistics.

\bibitem[{Chen et~al.(2012)Chen, Zhou, Zhu, and Xu}]{chen2012detecting}
Ying Chen, Yilu Zhou, Sencun Zhu, and Heng Xu. 2012.
\newblock Detecting offensive language in social media to protect adolescent
  online safety.
\newblock In \emph{2012 International Conference on Privacy, Security, Risk and
  Trust and 2012 International Confernece on Social Computing}, pages 71--80.
  IEEE.

\bibitem[{Chollampatt et~al.(2020)Chollampatt, Susanto, Tan, and
  Szymanska}]{chollampatt2020can}
Shamil Chollampatt, Raymond~Hendy Susanto, Liling Tan, and Ewa Szymanska. 2020.
\newblock Can automatic post-editing improve nmt?
\newblock \emph{arXiv preprint arXiv:2009.14395}.

\bibitem[{Chong and Druckman(2007)}]{chong2007framing}
Dennis Chong and James~N Druckman. 2007.
\newblock Framing theory.
\newblock \emph{Annual review of political science}, 10(1):103--126.

\bibitem[{Chowdhery et~al.(2022)Chowdhery, Narang, Devlin, Bosma, Mishra,
  Roberts, Barham, Chung, Sutton, Gehrmann, Schuh, Shi, Tsvyashchenko, Maynez,
  Rao, Barnes, Tay, Shazeer, Prabhakaran, Reif, Du, Hutchinson, Pope, Bradbury,
  Austin, Isard, Gur-Ari, Yin, Duke, Levskaya, Ghemawat, Dev, Michalewski,
  Garcia, Misra, Robinson, Fedus, Zhou, Ippolito, Luan, Lim, Zoph, Spiridonov,
  Sepassi, Dohan, Agrawal, Omernick, Dai, Pillai, Pellat, Lewkowycz, Moreira,
  Child, Polozov, Lee, Zhou, Wang, Saeta, Diaz, Firat, Catasta, Wei,
  Meier-Hellstern, Eck, Dean, Petrov, and
  Fiedel}]{https://doi.org/10.48550/arxiv.2204.02311}
Aakanksha Chowdhery, Sharan Narang, Jacob Devlin, Maarten Bosma, Gaurav Mishra,
  Adam Roberts, Paul Barham, Hyung~Won Chung, Charles Sutton, Sebastian
  Gehrmann, Parker Schuh, Kensen Shi, Sasha Tsvyashchenko, Joshua Maynez,
  Abhishek Rao, Parker Barnes, Yi~Tay, Noam Shazeer, Vinodkumar Prabhakaran,
  Emily Reif, Nan Du, Ben Hutchinson, Reiner Pope, James Bradbury, Jacob
  Austin, Michael Isard, Guy Gur-Ari, Pengcheng Yin, Toju Duke, Anselm
  Levskaya, Sanjay Ghemawat, Sunipa Dev, Henryk Michalewski, Xavier Garcia,
  Vedant Misra, Kevin Robinson, Liam Fedus, Denny Zhou, Daphne Ippolito, David
  Luan, Hyeontaek Lim, Barret Zoph, Alexander Spiridonov, Ryan Sepassi, David
  Dohan, Shivani Agrawal, Mark Omernick, Andrew~M. Dai,
  Thanumalayan~Sankaranarayana Pillai, Marie Pellat, Aitor Lewkowycz, Erica
  Moreira, Rewon Child, Oleksandr Polozov, Katherine Lee, Zongwei Zhou, Xuezhi
  Wang, Brennan Saeta, Mark Diaz, Orhan Firat, Michele Catasta, Jason Wei,
  Kathy Meier-Hellstern, Douglas Eck, Jeff Dean, Slav Petrov, and Noah Fiedel.
  2022.
\newblock \href {https://arxiv.org/abs/2204.02311} {Palm: Scaling language
  modeling with pathways}.

\bibitem[{Chronopoulou et~al.(2020)Chronopoulou, Stojanovski, and
  Fraser}]{chronopoulou-etal-2020-reusing}
Alexandra Chronopoulou, Dario Stojanovski, and Alexander Fraser. 2020.
\newblock \href {https://doi.org/10.18653/v1/2020.emnlp-main.214} {{R}eusing a
  {P}retrained {L}anguage {M}odel on {L}anguages with {L}imited {C}orpora for
  {U}nsupervised {NMT}}.
\newblock In \emph{Proceedings of the 2020 Conference on Empirical Methods in
  Natural Language Processing (EMNLP)}, pages 2703--2711, Online. Association
  for Computational Linguistics.

\bibitem[{Coates(2016)}]{genderCoatesJennifer2016WMaL}
Jennifer Coates. 2016.
\newblock \emph{Women, Men and Language: A Sociolinguistic Account of Gender
  Differences in Language}.
\newblock Routledge.

\bibitem[{Costa-juss{\`a} et~al.(2022)Costa-juss{\`a}, Smith, Ropers, Licht,
  Ferrando, and Escolano}]{costa2022toxicity}
Marta~R Costa-juss{\`a}, Eric Smith, Christophe Ropers, Daniel Licht, Javier
  Ferrando, and Carlos Escolano. 2022.
\newblock Toxicity in multilingual machine translation at scale.
\newblock \emph{arXiv preprint arXiv:2210.03070}.

\bibitem[{Crenshaw(2017)}]{kimberle2017}
Kimberlé~W. Crenshaw. 2017.
\newblock \emph{On Intersectionality: Essential Writings}.
\newblock The New Press, New York, NY.

\bibitem[{Da~San~Martino et~al.(2019)Da~San~Martino, Barr{\'o}n-Cede{\~n}o, and
  Nakov}]{da-san-martino-etal-2019-findings}
Giovanni Da~San~Martino, Alberto Barr{\'o}n-Cede{\~n}o, and Preslav Nakov.
  2019.
\newblock \href {https://doi.org/10.18653/v1/D19-5024} {Findings of the
  {NLP}4{IF}-2019 shared task on fine-grained propaganda detection}.
\newblock In \emph{Proceedings of the Second Workshop on Natural Language
  Processing for Internet Freedom: Censorship, Disinformation, and Propaganda},
  pages 162--170, Hong Kong, China. Association for Computational Linguistics.

\bibitem[{Da~San~Martino et~al.(2020)Da~San~Martino, Barr{\'o}n-Cede{\~n}o,
  Wachsmuth, Petrov, and Nakov}]{da-san-martino-etal-2020-semeval}
Giovanni Da~San~Martino, Alberto Barr{\'o}n-Cede{\~n}o, Henning Wachsmuth,
  Rostislav Petrov, and Preslav Nakov. 2020.
\newblock \href {https://doi.org/10.18653/v1/2020.semeval-1.186}
  {{S}em{E}val-2020 task 11: Detection of propaganda techniques in news
  articles}.
\newblock In \emph{Proceedings of the Fourteenth Workshop on Semantic
  Evaluation}, pages 1377--1414, Barcelona (online). International Committee
  for Computational Linguistics.

\bibitem[{Dadvar et~al.(2012)Dadvar, Jong, Ordelman, and
  Trieschnigg}]{dadvar2012improved}
Maral Dadvar, FMG~de Jong, Roeland Ordelman, and Dolf Trieschnigg. 2012.
\newblock Improved cyberbullying detection using gender information.
\newblock In \emph{Proceedings of the Twelfth Dutch-Belgian Information
  Retrieval Workshop (DIR 2012)}. University of Ghent.

\bibitem[{Dathathri et~al.(2019)Dathathri, Madotto, Lan, Hung, Frank, Molino,
  Yosinski, and Liu}]{dathathri2019plug}
Sumanth Dathathri, Andrea Madotto, Janice Lan, Jane Hung, Eric Frank, Piero
  Molino, Jason Yosinski, and Rosanne Liu. 2019.
\newblock Plug and play language models: A simple approach to controlled text
  generation.
\newblock \emph{ICLR}.

\bibitem[{Davidson et~al.(2017{\natexlab{a}})Davidson, Warmsley, Macy, and
  Weber}]{davidson2017automated}
Thomas Davidson, Dana Warmsley, Michael Macy, and Ingmar Weber.
  2017{\natexlab{a}}.
\newblock Automated hate speech detection and the problem of offensive
  language.
\newblock In \emph{Proceedings of the International AAAI Conference on Web and
  Social Media}, volume~11.

\bibitem[{Davidson et~al.(2017{\natexlab{b}})Davidson, Warmsley, Macy, and
  Weber}]{Davidson2017AutomatedHS}
Thomas Davidson, Dana Warmsley, Michael~W. Macy, and Ingmar Weber.
  2017{\natexlab{b}}.
\newblock Automated hate speech detection and the problem of offensive
  language.
\newblock In \emph{ICWSM}.

\bibitem[{Daws(2020)}]{gpt3}
Ryan Daws. 2020.
\newblock \href
  {https://www.artificialintelligence-news.com/2020/10/28/medical-chatbot-openai-gpt3-patient-kill-themselves/r}
  {{Medical chatbot using OpenAI’s GPT-3 told a fake patient to kill
  themselves}}.

\bibitem[{De~Cao et~al.(2021)De~Cao, Aziz, and
  Titov}]{de-cao-etal-2021-editing}
Nicola De~Cao, Wilker Aziz, and Ivan Titov. 2021.
\newblock \href {https://doi.org/10.18653/v1/2021.emnlp-main.522} {Editing
  factual knowledge in language models}.
\newblock In \emph{Proceedings of the 2021 Conference on Empirical Methods in
  Natural Language Processing}, pages 6491--6506, Online and Punta Cana,
  Dominican Republic. Association for Computational Linguistics.

\bibitem[{de~Masson~d\textquotesingle Autume
  et~al.(2019)de~Masson~d\textquotesingle Autume, Ruder, Kong, and
  Yogatama}]{NEURIPS2019_f8d2e80c}
Cyprien de~Masson~d\textquotesingle Autume, Sebastian Ruder, Lingpeng Kong, and
  Dani Yogatama. 2019.
\newblock \href
  {https://proceedings.neurips.cc/paper/2019/file/f8d2e80c1458ea2501f98a2cafadb397-Paper.pdf}
  {Episodic memory in lifelong language learning}.
\newblock In \emph{Advances in Neural Information Processing Systems},
  volume~32. Curran Associates, Inc.

\bibitem[{Denton et~al.(2020)Denton, Hanna, Amironesei, Smart, Nicole, and
  Scheuerman}]{denton2020bringing}
Emily Denton, Alex Hanna, Razvan Amironesei, Andrew Smart, Hilary Nicole, and
  Morgan~Klaus Scheuerman. 2020.
\newblock Bringing the people back in: Contesting benchmark machine learning
  datasets.
\newblock \emph{arXiv preprint arXiv:2007.07399}.

\bibitem[{Devlin et~al.(2018)Devlin, Chang, Lee, and
  Toutanova}]{devlin2018bert}
Jacob Devlin, Ming-Wei Chang, Kenton Lee, and Kristina Toutanova. 2018.
\newblock Bert: Pre-training of deep bidirectional transformers for language
  understanding.
\newblock \emph{arXiv preprint arXiv:1810.04805}.

\bibitem[{Dhingra et~al.(2022)Dhingra, Cole, Eisenschlos, Gillick, Eisenstein,
  and Cohen}]{dhingra-etal-2022-time}
Bhuwan Dhingra, Jeremy~R. Cole, Julian~Martin Eisenschlos, Daniel Gillick,
  Jacob Eisenstein, and William~W. Cohen. 2022.
\newblock \href {https://doi.org/10.1162/tacl_a_00459} {Time-aware language
  models as temporal knowledge bases}.
\newblock \emph{Transactions of the Association for Computational Linguistics},
  10:257--273.

\bibitem[{Dinan et~al.(2020)Dinan, Fan, Williams, Urbanek, Kiela, and
  Weston}]{dinan-etal-2020-queens}
Emily Dinan, Angela Fan, Adina Williams, Jack Urbanek, Douwe Kiela, and Jason
  Weston. 2020.
\newblock \href {https://doi.org/10.18653/v1/2020.emnlp-main.656} {Queens are
  powerful too: Mitigating gender bias in dialogue generation}.
\newblock In \emph{Proceedings of the 2020 Conference on Empirical Methods in
  Natural Language Processing (EMNLP)}, pages 8173--8188, Online. Association
  for Computational Linguistics.

\bibitem[{Dinan et~al.(2019)Dinan, Roller, Shuster, Fan, Auli, and
  Weston}]{dinan2018wizard}
Emily Dinan, Stephen Roller, Kurt Shuster, Angela Fan, Michael Auli, and Jason
  Weston. 2019.
\newblock \href {https://openreview.net/forum?id=r1l73iRqKm} {Wizard of
  wikipedia: Knowledge-powered conversational agents}.
\newblock In \emph{International Conference on Learning Representations}.

\bibitem[{Dixon et~al.(2018)Dixon, Li, Sorensen, Thain, and
  Vasserman}]{dixon2018measuring}
Lucas Dixon, John Li, Jeffrey Sorensen, Nithum Thain, and Lucy Vasserman. 2018.
\newblock Measuring and mitigating unintended bias in text classification.
\newblock In \emph{Proceedings of the 2018 AAAI/ACM Conference on AI, Ethics,
  and Society}, pages 67--73.

\bibitem[{Dodge et~al.(2021)Dodge, Sap, Marasovi{\'c}, Agnew, Ilharco,
  Groeneveld, Mitchell, and Gardner}]{dodge2021documenting}
Jesse Dodge, Maarten Sap, Ana Marasovi{\'c}, William Agnew, Gabriel Ilharco,
  Dirk Groeneveld, Margaret Mitchell, and Matt Gardner. 2021.
\newblock Documenting large webtext corpora: A case study on the colossal clean
  crawled corpus.
\newblock \emph{arXiv preprint arXiv:2104.08758}.

\bibitem[{d'Sa et~al.(2020)d'Sa, Illina, and Fohr}]{d2020bert}
Ashwin~Geet d'Sa, Irina Illina, and Dominique Fohr. 2020.
\newblock Bert and fasttext embeddings for automatic detection of toxic speech.
\newblock In \emph{2020 International Multi-Conference on:“Organization of
  Knowledge and Advanced Technologies”(OCTA)}, pages 1--5. IEEE.

\bibitem[{Dugan et~al.(2020)Dugan, Ippolito, Kirubarajan, and
  Callison-Burch}]{dugan-etal-2020-roft}
Liam Dugan, Daphne Ippolito, Arun Kirubarajan, and Chris Callison-Burch. 2020.
\newblock \href {https://doi.org/10.18653/v1/2020.emnlp-demos.25} {{R}o{FT}: A
  tool for evaluating human detection of machine-generated text}.
\newblock In \emph{Proceedings of the 2020 Conference on Empirical Methods in
  Natural Language Processing: System Demonstrations}, pages 189--196, Online.
  Association for Computational Linguistics.

\bibitem[{Eckert and McConnell-Ginet(2003)}]{gendereckert_mcconnell-ginet_2003}
Penelope Eckert and Sally McConnell-Ginet. 2003.
\newblock \emph{Language and Gender}.
\newblock Cambridge University Press.

\bibitem[{Falke et~al.(2019)Falke, Ribeiro, Utama, Dagan, and
  Gurevych}]{DBLP:conf/acl/FalkeRUDG19}
Tobias Falke, Leonardo F.~R. Ribeiro, Prasetya~Ajie Utama, Ido Dagan, and Iryna
  Gurevych. 2019.
\newblock \href {https://doi.org/10.18653/v1/p19-1213} {Ranking generated
  summaries by correctness: An interesting but challenging application for
  natural language inference}.
\newblock In \emph{ACL (1)}, pages 2214--2220.

\bibitem[{Fan et~al.(2021)Fan, Gardent, Braud, and
  Bordes}]{10.1162/tacl_a_00356}
Angela Fan, Claire Gardent, Chloé Braud, and Antoine Bordes. 2021.
\newblock \href {https://doi.org/10.1162/tacl_a_00356} {{Augmenting
  Transformers with KNN-Based Composite Memory for Dialog}}.
\newblock \emph{Transactions of the Association for Computational Linguistics},
  9:82--99.

\bibitem[{Fan et~al.(2018)Fan, Lewis, and Dauphin}]{fan-etal-2018-hierarchical}
Angela Fan, Mike Lewis, and Yann Dauphin. 2018.
\newblock \href {https://doi.org/10.18653/v1/P18-1082} {Hierarchical neural
  story generation}.
\newblock In \emph{Proceedings of the 56th Annual Meeting of the Association
  for Computational Linguistics (Volume 1: Long Papers)}, pages 889--898,
  Melbourne, Australia. Association for Computational Linguistics.

\bibitem[{Feldman et~al.(2021)Feldman, Da~San~Martino, Leberknight, and
  Nakov}]{feldman2021proceedings}
Anna Feldman, Giovanni Da~San~Martino, Chris Leberknight, and Preslav Nakov.
  2021.
\newblock Proceedings of the fourth workshop on nlp for internet freedom:
  Censorship, disinformation, and propaganda.
\newblock In \emph{Proceedings of the Fourth Workshop on NLP for Internet
  Freedom: Censorship, Disinformation, and Propaganda}.

\bibitem[{Field et~al.(2021)Field, Blodgett, Waseem, and
  Tsvetkov}]{field-etal-2021-survey}
Anjalie Field, Su~Lin Blodgett, Zeerak Waseem, and Yulia Tsvetkov. 2021.
\newblock \href {https://doi.org/10.18653/v1/2021.acl-long.149} {A survey of
  race, racism, and anti-racism in {NLP}}.
\newblock In \emph{Proceedings of the 59th Annual Meeting of the Association
  for Computational Linguistics and the 11th International Joint Conference on
  Natural Language Processing (Volume 1: Long Papers)}, pages 1905--1925,
  Online. Association for Computational Linguistics.

\bibitem[{Field and Tsvetkov(2020)}]{Field2020UnsupervisedDO}
Anjalie Field and Yulia Tsvetkov. 2020.
\newblock Unsupervised discovery of implicit gender bias.
\newblock In \emph{Proceedings of the 2020 Conference on Empirical Methods in
  Natural Language Processing (EMNLP)}, pages 596--608.

\bibitem[{Gamb{\"a}ck and Sikdar(2017)}]{gamback2017using}
Bj{\"o}rn Gamb{\"a}ck and Utpal~Kumar Sikdar. 2017.
\newblock Using convolutional neural networks to classify hate-speech.
\newblock In \emph{Proceedings of the first workshop on abusive language
  online}, pages 85--90.

\bibitem[{Gehman et~al.(2020)Gehman, Gururangan, Sap, Choi, and
  Smith}]{gehman-etal-2020-realtoxicityprompts}
Samuel Gehman, Suchin Gururangan, Maarten Sap, Yejin Choi, and Noah~A. Smith.
  2020.
\newblock \href {https://doi.org/10.18653/v1/2020.findings-emnlp.301}
  {{R}eal{T}oxicity{P}rompts: Evaluating neural toxic degeneration in language
  models}.
\newblock In \emph{Findings of the Association for Computational Linguistics:
  EMNLP 2020}, pages 3356--3369, Online. Association for Computational
  Linguistics.

\bibitem[{Gehrmann et~al.(2019)Gehrmann, Strobelt, and
  Rush}]{gehrmann-etal-2019-gltr}
Sebastian Gehrmann, Hendrik Strobelt, and Alexander Rush. 2019.
\newblock \href {https://doi.org/10.18653/v1/P19-3019} {{GLTR}: Statistical
  detection and visualization of generated text}.
\newblock In \emph{Proceedings of the 57th Annual Meeting of the Association
  for Computational Linguistics: System Demonstrations}, pages 111--116,
  Florence, Italy. Association for Computational Linguistics.

\bibitem[{Geva et~al.(2022)Geva, Caciularu, Wang, and
  Goldberg}]{Geva2022TransformerFL}
Mor Geva, Avi Caciularu, Ke~Wang, and Yoav Goldberg. 2022.
\newblock Transformer feed-forward layers build predictions by promoting
  concepts in the vocabulary space.
\newblock \emph{ArXiv}, abs/2203.14680.

\bibitem[{Geva et~al.(2019)Geva, Goldberg, and Berant}]{geva2019we}
Mor Geva, Yoav Goldberg, and Jonathan Berant. 2019.
\newblock Are we modeling the task or the annotator? an investigation of
  annotator bias in natural language understanding datasets.
\newblock In \emph{Proceedings of the 2019 Conference on Empirical Methods in
  Natural Language Processing and the 9th International Joint Conference on
  Natural Language Processing (EMNLP-IJCNLP)}, pages 1161--1166.

\bibitem[{Gleason(2022)}]{chatgpt-education2}
Nancy Gleason. 2022.
\newblock \href
  {https://www.timeshighereducation.com/campus/chatgpt-and-rise-ai-writers-how-should-higher-education-respond}
  {Chatgpt and the rise of ai writers: how should higher education respond?}

\bibitem[{Goyal and Durrett(2020)}]{dae}
Tanya Goyal and Greg Durrett. 2020.
\newblock \href {https://doi.org/10.18653/v1/2020.findings-emnlp.322}
  {Evaluating factuality in generation with dependency-level entailment}.
\newblock In \emph{Findings of the Association for Computational Linguistics:
  EMNLP 2020}, pages 3592--3603, Online. Association for Computational
  Linguistics.

\bibitem[{Graves(2012)}]{graves2012sequence}
Alex Graves. 2012.
\newblock Sequence transduction with recurrent neural networks.
\newblock \emph{arXiv preprint arXiv:1211.3711}.

\bibitem[{Guan et~al.(2020)Guan, Huang, Zhao, Zhu, and
  Huang}]{guan2020knowledge}
Jian Guan, Fei Huang, Zhihao Zhao, Xiaoyan Zhu, and Minlie Huang. 2020.
\newblock A knowledge-enhanced pretraining model for commonsense story
  generation.
\newblock \emph{Transactions of the Association for Computational Linguistics},
  8:93--108.

\bibitem[{Guo et~al.(2022)Guo, Schlichtkrull, and
  Vlachos}]{guo-etal-2022-survey}
Zhijiang Guo, Michael Schlichtkrull, and Andreas Vlachos. 2022.
\newblock \href {https://doi.org/10.1162/tacl_a_00454} {A survey on automated
  fact-checking}.
\newblock \emph{Transactions of the Association for Computational Linguistics},
  10:178--206.

\bibitem[{Gururangan et~al.(2020)Gururangan, Marasovi{\'c}, Swayamdipta, Lo,
  Beltagy, Downey, and Smith}]{gururangan-etal-2020-dont}
Suchin Gururangan, Ana Marasovi{\'c}, Swabha Swayamdipta, Kyle Lo, Iz~Beltagy,
  Doug Downey, and Noah~A. Smith. 2020.
\newblock \href {https://doi.org/10.18653/v1/2020.acl-main.740} {Don{'}t stop
  pretraining: Adapt language models to domains and tasks}.
\newblock In \emph{Proceedings of the 58th Annual Meeting of the Association
  for Computational Linguistics}, pages 8342--8360, Online. Association for
  Computational Linguistics.

\bibitem[{Gururangan et~al.(2018)Gururangan, Swayamdipta, Levy, Schwartz,
  Bowman, and Smith}]{gururangan-etal-2018-annotation}
Suchin Gururangan, Swabha Swayamdipta, Omer Levy, Roy Schwartz, Samuel Bowman,
  and Noah~A. Smith. 2018.
\newblock \href {https://doi.org/10.18653/v1/N18-2017} {Annotation artifacts in
  natural language inference data}.
\newblock In \emph{Proceedings of the 2018 Conference of the North {A}merican
  Chapter of the Association for Computational Linguistics: Human Language
  Technologies, Volume 2 (Short Papers)}, pages 107--112, New Orleans,
  Louisiana. Association for Computational Linguistics.

\bibitem[{Hall et~al.(2022)Hall, van~der Maaten, Gustafson, and
  Adcock}]{Hall2022ASS}
Melissa~R.H. Hall, Laurens van~der Maaten, Laura Gustafson, and Aaron~B.
  Adcock. 2022.
\newblock A systematic study of bias amplification.
\newblock \emph{arXiv preprint arXiv:2301.11305}.

\bibitem[{Han and Tsvetkov(2020)}]{han-tsvetkov-2020-fortifying}
Xiaochuang Han and Yulia Tsvetkov. 2020.
\newblock \href {https://doi.org/10.18653/v1/2020.emnlp-main.622} {Fortifying
  toxic speech detectors against veiled toxicity}.
\newblock In \emph{Proceedings of the 2020 Conference on Empirical Methods in
  Natural Language Processing (EMNLP)}, pages 7732--7739, Online. Association
  for Computational Linguistics.

\bibitem[{Han and Tsvetkov(2021)}]{han-tsvetkov-2021-influence-tuning}
Xiaochuang Han and Yulia Tsvetkov. 2021.
\newblock \href {https://doi.org/10.18653/v1/2021.findings-emnlp.374}
  {Influence tuning: Demoting spurious correlations via instance attribution
  and instance-driven updates}.
\newblock In \emph{Findings of the Association for Computational Linguistics:
  EMNLP 2021}, pages 4398--4409, Punta Cana, Dominican Republic. Association
  for Computational Linguistics.

\bibitem[{He et~al.(2021{\natexlab{a}})He, Neubig, and
  Berg-Kirkpatrick}]{he-etal-2021-efficient}
Junxian He, Graham Neubig, and Taylor Berg-Kirkpatrick. 2021{\natexlab{a}}.
\newblock \href {https://doi.org/10.18653/v1/2021.emnlp-main.461} {Efficient
  nearest neighbor language models}.
\newblock In \emph{Proceedings of the 2021 Conference on Empirical Methods in
  Natural Language Processing}, pages 5703--5714, Online and Punta Cana,
  Dominican Republic. Association for Computational Linguistics.

\bibitem[{He et~al.(2021{\natexlab{b}})He, Majumder, and
  McAuley}]{he-etal-2021-detect-perturb}
Zexue He, Bodhisattwa~Prasad Majumder, and Julian McAuley. 2021{\natexlab{b}}.
\newblock \href {https://doi.org/10.18653/v1/2021.findings-emnlp.352} {Detect
  and perturb: Neutral rewriting of biased and sensitive text via
  gradient-based decoding}.
\newblock In \emph{Findings of the Association for Computational Linguistics:
  EMNLP 2021}, pages 4173--4181, Punta Cana, Dominican Republic. Association
  for Computational Linguistics.

\bibitem[{Henderson et~al.(2022)Henderson, Krass, Zheng, Guha, Manning,
  Jurafsky, and Ho}]{Henderson2022PileOL}
Peter Henderson, Mark~S. Krass, Lucia Zheng, Neel Guha, Christopher~D. Manning,
  Dan Jurafsky, and Daniel~E. Ho. 2022.
\newblock Pile of law: Learning responsible data filtering from the law and a
  256gb open-source legal dataset.
\newblock \emph{NeurIPS}.

\bibitem[{Hendrycks et~al.(2021)Hendrycks, Burns, Basart, Critch, Li, Song, and
  Steinhardt}]{hendrycks2021ethics}
Dan Hendrycks, Collin Burns, Steven Basart, Andrew Critch, Jerry Li, Dawn Song,
  and Jacob Steinhardt. 2021.
\newblock Aligning ai with shared human values.
\newblock \emph{Proceedings of the International Conference on Learning
  Representations (ICLR)}.

\bibitem[{Hershcovich et~al.(2022)Hershcovich, Frank, Lent, de~Lhoneux, Abdou,
  Brandl, Bugliarello, Cabello~Piqueras, Chalkidis, Cui, Fierro, Margatina,
  Rust, and S{\o}gaard}]{hershcovich-etal-2022-challenges}
Daniel Hershcovich, Stella Frank, Heather Lent, Miryam de~Lhoneux, Mostafa
  Abdou, Stephanie Brandl, Emanuele Bugliarello, Laura Cabello~Piqueras, Ilias
  Chalkidis, Ruixiang Cui, Constanza Fierro, Katerina Margatina, Phillip Rust,
  and Anders S{\o}gaard. 2022.
\newblock \href {https://doi.org/10.18653/v1/2022.acl-long.482} {Challenges and
  strategies in cross-cultural {NLP}}.
\newblock In \emph{Proceedings of the 60th Annual Meeting of the Association
  for Computational Linguistics (Volume 1: Long Papers)}, pages 6997--7013,
  Dublin, Ireland. Association for Computational Linguistics.

\bibitem[{Hoang et~al.(2017)Hoang, Haffari, and Cohn}]{hoang-etal-2017-towards}
Cong Duy~Vu Hoang, Gholamreza Haffari, and Trevor Cohn. 2017.
\newblock \href {https://doi.org/10.18653/v1/D17-1014} {Towards decoding as
  continuous optimisation in neural machine translation}.
\newblock In \emph{Proceedings of the 2017 Conference on Empirical Methods in
  Natural Language Processing}, pages 146--156, Copenhagen, Denmark.
  Association for Computational Linguistics.

\bibitem[{Holmes and Wilson(2017)}]{holmes2017introduction}
Janet Holmes and Nick Wilson. 2017.
\newblock \emph{An introduction to sociolinguistics}.
\newblock Routledge.

\bibitem[{Holtzman et~al.(2020)Holtzman, Buys, Du, Forbes, and
  Choi}]{Holtzman2020The}
Ari Holtzman, Jan Buys, Li~Du, Maxwell Forbes, and Yejin Choi. 2020.
\newblock \href {https://openreview.net/forum?id=rygGQyrFvH} {The curious case
  of neural text degeneration}.
\newblock In \emph{International Conference on Learning Representations}.

\bibitem[{Hossain et~al.(2020)Hossain, Ghazvininejad, and
  Zettlemoyer}]{hossain-etal-2020-simple}
Nabil Hossain, Marjan Ghazvininejad, and Luke Zettlemoyer. 2020.
\newblock \href {https://doi.org/10.18653/v1/2020.acl-main.228} {Simple and
  effective retrieve-edit-rerank text generation}.
\newblock In \emph{Proceedings of the 58th Annual Meeting of the Association
  for Computational Linguistics}, pages 2532--2538, Online. Association for
  Computational Linguistics.

\bibitem[{Huang et~al.(2022)Huang, Nakov, Choi, and Ji}]{Huang2022FakingFN}
Kung-Hsiang Huang, Preslav Nakov, Yejin Choi, and Heng Ji. 2022.
\newblock Faking fake news for real fake news detection: Propaganda-loaded
  training data generation.
\newblock \emph{ArXiv}, abs/2203.05386.

\bibitem[{Huang et~al.(2020)Huang, Wu, and Wang}]{huang-etal-2020-knowledge}
Luyang Huang, Lingfei Wu, and Lu~Wang. 2020.
\newblock \href {https://doi.org/10.18653/v1/2020.acl-main.457} {Knowledge
  graph-augmented abstractive summarization with semantic-driven cloze reward}.
\newblock In \emph{Proceedings of the 58th Annual Meeting of the Association
  for Computational Linguistics}, pages 5094--5107, Online. Association for
  Computational Linguistics.

\bibitem[{Hunt(2016)}]{tay}
Elle Hunt. 2016.
\newblock \href
  {https://www.theguardian.com/technology/2016/mar/24/tay-microsofts-ai-chatbot-gets-a-crash-course-in-racism-from-twitter}
  {{Tay, Microsoft's AI chatbot, gets a crash course in racism from Twitter}}.

\bibitem[{Hutchinson et~al.(2021)Hutchinson, Smart, Hanna, Denton, Greer,
  Kjartansson, Barnes, and Mitchell}]{10.1145/3442188.3445918}
Ben Hutchinson, Andrew Smart, Alex Hanna, Emily Denton, Christina Greer, Oddur
  Kjartansson, Parker Barnes, and Margaret Mitchell. 2021.
\newblock \href {https://doi.org/10.1145/3442188.3445918} {Towards
  accountability for machine learning datasets: Practices from software
  engineering and infrastructure}.
\newblock In \emph{Proceedings of the 2021 ACM Conference on Fairness,
  Accountability, and Transparency}, FAccT '21, page 560–575, New York, NY,
  USA. Association for Computing Machinery.

\bibitem[{Ippolito et~al.(2020)Ippolito, Duckworth, Callison-Burch, and
  Eck}]{ippolito-etal-2020-automatic}
Daphne Ippolito, Daniel Duckworth, Chris Callison-Burch, and Douglas Eck. 2020.
\newblock \href {https://doi.org/10.18653/v1/2020.acl-main.164} {Automatic
  detection of generated text is easiest when humans are fooled}.
\newblock In \emph{Proceedings of the 58th Annual Meeting of the Association
  for Computational Linguistics}, pages 1808--1822, Online. Association for
  Computational Linguistics.

\bibitem[{Ippolito et~al.(2022)Ippolito, Tram{\`e}r, Nasr, Zhang, Jagielski,
  Lee, Choquette-Choo, and Carlini}]{Ippolito2022PreventingVM}
Daphne Ippolito, Florian Tram{\`e}r, Milad Nasr, Chiyuan Zhang, Matthew
  Jagielski, Katherine Lee, Christopher~A. Choquette-Choo, and Nicholas
  Carlini. 2022.
\newblock Preventing verbatim memorization in language models gives a false
  sense of privacy.
\newblock \emph{ArXiv}, abs/2210.17546.

\bibitem[{Izacard and Grave(2021)}]{izacard-grave-2021-leveraging}
Gautier Izacard and Edouard Grave. 2021.
\newblock \href {https://doi.org/10.18653/v1/2021.eacl-main.74} {Leveraging
  passage retrieval with generative models for open domain question answering}.
\newblock In \emph{Proceedings of the 16th Conference of the European Chapter
  of the Association for Computational Linguistics: Main Volume}, pages
  874--880, Online. Association for Computational Linguistics.

\bibitem[{Jacovi et~al.(2021)Jacovi, Marasovi{\'c}, Miller, and
  Goldberg}]{Jacovi2021FormalizingTI}
Alon Jacovi, Ana Marasovi{\'c}, Tim Miller, and Yoav Goldberg. 2021.
\newblock Formalizing trust in artificial intelligence: Prerequisites, causes
  and goals of human trust in ai.
\newblock \emph{Proc. FAccT}.

\bibitem[{Jang(2021)}]{racict-korean}
Heesoo Jang. 2021.
\newblock \href
  {https://slate.com/technology/2021/04/scatterlab-lee-luda-chatbot-kakaotalk-ai-privacy.html}
  {A {South Korean} chatbot shows just how sloppy tech companies can be with
  user data}.

\bibitem[{Jang et~al.(2022)Jang, Yoon, Yang, Cha, Lee, Logeswaran, and
  Seo}]{Jang2022KnowledgeUF}
Joel Jang, Dongkeun Yoon, Sohee Yang, Sungmin Cha, Moontae Lee, Lajanugen
  Logeswaran, and Minjoon Seo. 2022.
\newblock Knowledge unlearning for mitigating privacy risks in language models.
\newblock \emph{ArXiv}, abs/2210.01504.

\bibitem[{Jawahar et~al.(2020)Jawahar, Abdul-Mageed, and
  Lakshmanan}]{jawahar-etal-2020-automatic}
Ganesh Jawahar, Muhammad Abdul-Mageed, and Laks Lakshmanan, V.S. 2020.
\newblock \href {https://doi.org/10.18653/v1/2020.coling-main.208} {Automatic
  detection of machine generated text: A critical survey}.
\newblock In \emph{Proceedings of the 28th International Conference on
  Computational Linguistics}, pages 2296--2309, Barcelona, Spain (Online).
  International Committee on Computational Linguistics.

\bibitem[{Ji et~al.(2020)Ji, Ke, Huang, Wei, Zhu, and
  Huang}]{ji-etal-2020-language}
Haozhe Ji, Pei Ke, Shaohan Huang, Furu Wei, Xiaoyan Zhu, and Minlie Huang.
  2020.
\newblock \href {https://doi.org/10.18653/v1/2020.emnlp-main.54} {Language
  generation with multi-hop reasoning on commonsense knowledge graph}.
\newblock In \emph{Proceedings of the 2020 Conference on Empirical Methods in
  Natural Language Processing (EMNLP)}, pages 725--736, Online. Association for
  Computational Linguistics.

\bibitem[{Jiang et~al.(2021)Jiang, Hwang, Bhagavatula, Bras, Forbes, Borchardt,
  Liang, Etzioni, Sap, and Choi}]{Jiang2021DelphiTM}
Liwei Jiang, Jena~D. Hwang, Chandrasekhar Bhagavatula, Ronan~Le Bras, Maxwell
  Forbes, Jon Borchardt, Jenny Liang, Oren Etzioni, Maarten Sap, and Yejin
  Choi. 2021.
\newblock Delphi: Towards machine ethics and norms.
\newblock \emph{ArXiv}, abs/2110.07574.

\bibitem[{Jo and Gebru(2020)}]{10.1145/3351095.3372829}
Eun~Seo Jo and Timnit Gebru. 2020.
\newblock \href {https://doi.org/10.1145/3351095.3372829} {Lessons from
  archives: Strategies for collecting sociocultural data in machine learning}.
\newblock In \emph{Proceedings of the 2020 Conference on Fairness,
  Accountability, and Transparency}, FAT* '20, page 306–316, New York, NY,
  USA. Association for Computing Machinery.

\bibitem[{Joshi et~al.(2020)Joshi, Santy, Budhiraja, Bali, and
  Choudhury}]{joshi-etal-2020-state}
Pratik Joshi, Sebastin Santy, Amar Budhiraja, Kalika Bali, and Monojit
  Choudhury. 2020.
\newblock \href {https://doi.org/10.18653/v1/2020.acl-main.560} {The state and
  fate of linguistic diversity and inclusion in the {NLP} world}.
\newblock In \emph{Proceedings of the 58th Annual Meeting of the Association
  for Computational Linguistics}, pages 6282--6293, Online. Association for
  Computational Linguistics.

\bibitem[{Kammoun et~al.(2022)Kammoun, Slama, Tabia, Ouni, and
  Abid}]{Kammoun_2022}
Amina Kammoun, Rim Slama, Hedi Tabia, Tarek Ouni, and Mohmed Abid. 2022.
\newblock \href {https://doi.org/10.1145/1122445.1122456} {Generative
  adversarial networks for face generation: A survey}.
\newblock \emph{{ACM} Computing Surveys}.

\bibitem[{Kandpal et~al.(2022)Kandpal, Wallace, and
  Raffel}]{Kandpal2022DeduplicatingTD}
Nikhil Kandpal, Eric Wallace, and Colin Raffel. 2022.
\newblock Deduplicating training data mitigates privacy risks in language
  models.
\newblock \emph{ICML}.

\bibitem[{Kerrigan et~al.(2020)Kerrigan, Slack, and
  Tuyls}]{kerrigan-etal-2020-differentially}
Gavin Kerrigan, Dylan Slack, and Jens Tuyls. 2020.
\newblock \href {https://doi.org/10.18653/v1/2020.privatenlp-1.5}
  {Differentially private language models benefit from public pre-training}.
\newblock In \emph{Proceedings of the Second Workshop on Privacy in NLP}, pages
  39--45, Online. Association for Computational Linguistics.

\bibitem[{Keskar et~al.(2019)Keskar, McCann, Varshney, Xiong, and
  Socher}]{keskarCTRL2019}
Nitish~Shirish Keskar, Bryan McCann, Lav Varshney, Caiming Xiong, and Richard
  Socher. 2019.
\newblock {CTRL - A Conditional Transformer Language Model for Controllable
  Generation}.
\newblock \emph{ACM Computing}.

\bibitem[{Khandelwal et~al.(2020)Khandelwal, Levy, Jurafsky, Zettlemoyer, and
  Lewis}]{Khandelwal2020Generalization}
Urvashi Khandelwal, Omer Levy, Dan Jurafsky, Luke Zettlemoyer, and Mike Lewis.
  2020.
\newblock \href {https://openreview.net/forum?id=HklBjCEKvH} {Generalization
  through memorization: Nearest neighbor language models}.
\newblock In \emph{International Conference on Learning Representations}.

\bibitem[{Kim(2016)}]{privacy2016kim}
Dongwoo Kim. 2016.
\newblock \href
  {https://thediplomat.com/2021/01/chatbot-gone-awry-starts-conversations-about-ai-ethics-in-south-korea/}
  {Chatbot gone awry starts conversations about ai ethics in south korea}.

\bibitem[{King et~al.(2022)King, Shen, Subramani, Weld, Beltagy, and
  Downey}]{king2022don}
Daniel King, Zejiang Shen, Nishant Subramani, Daniel~S Weld, Iz~Beltagy, and
  Doug Downey. 2022.
\newblock Don't say what you don't know: Improving the consistency of
  abstractive summarization by constraining beam search.
\newblock \emph{arXiv preprint arXiv:2203.08436}.

\bibitem[{Koenecke et~al.(2020)Koenecke, Nam, Lake, Nudell, Quartey, Mengesha,
  Toups, Rickford, Jurafsky, and Goel}]{doi:10.1073/pnas.1915768117}
Allison Koenecke, Andrew Nam, Emily Lake, Joe Nudell, Minnie Quartey, Zion
  Mengesha, Connor Toups, John~R. Rickford, Dan Jurafsky, and Sharad Goel.
  2020.
\newblock Racial disparities in automated speech recognition.
\newblock \emph{Proceedings of the National Academy of Sciences}.

\bibitem[{Koh and Liang(2017)}]{pmlr-v70-koh17a}
Pang~Wei Koh and Percy Liang. 2017.
\newblock \href {https://proceedings.mlr.press/v70/koh17a.html} {Understanding
  black-box predictions via influence functions}.
\newblock In \emph{Proceedings of the 34th International Conference on Machine
  Learning}, volume~70 of \emph{Proceedings of Machine Learning Research},
  pages 1885--1894. PMLR.

\bibitem[{Korzeniowski et~al.(2019)Korzeniowski, Rolczynski, Sadownik, Korbak,
  and Możejko}]{Korzeniowski2019ExploitingUP}
Renard Korzeniowski, Rafal Rolczynski, Przemyslaw Sadownik, Tomasz Korbak, and
  Marcin Możejko. 2019.
\newblock Exploiting unsupervised pre-training and automated feature
  engineering for low-resource hate speech detection in polish.
\newblock \emph{ArXiv}, abs/1906.09325.

\bibitem[{Krause et~al.(2021)Krause, Gotmare, McCann, Keskar, Joty, Socher, and
  Rajani}]{krause-etal-2021-gedi-generative}
Ben Krause, Akhilesh~Deepak Gotmare, Bryan McCann, Nitish~Shirish Keskar,
  Shafiq Joty, Richard Socher, and Nazneen~Fatema Rajani. 2021.
\newblock \href {https://doi.org/10.18653/v1/2021.findings-emnlp.424}
  {{G}e{D}i: Generative discriminator guided sequence generation}.
\newblock In \emph{Findings of the Association for Computational Linguistics:
  EMNLP 2021}, pages 4929--4952, Punta Cana, Dominican Republic. Association
  for Computational Linguistics.

\bibitem[{Krishna et~al.(2022)Krishna, yin Chang, Wieting, and
  Iyyer}]{Krishna2022RankGenIT}
Kalpesh Krishna, Ya~yin Chang, John Wieting, and Mohit Iyyer. 2022.
\newblock Rankgen: Improving text generation with large ranking models.
\newblock \emph{EMNLP}.

\bibitem[{Kryscinski et~al.(2020)Kryscinski, McCann, Xiong, and
  Socher}]{factcc}
Wojciech Kryscinski, Bryan McCann, Caiming Xiong, and Richard Socher. 2020.
\newblock \href {https://doi.org/10.18653/v1/2020.emnlp-main.750} {Evaluating
  the factual consistency of abstractive text summarization}.
\newblock In \emph{Proceedings of the 2020 Conference on Empirical Methods in
  Natural Language Processing (EMNLP)}, pages 9332--9346, Online. Association
  for Computational Linguistics.

\bibitem[{Kumar et~al.(2021{\natexlab{a}})Kumar, Anastasopoulos, Wintner, and
  Tsvetkov}]{kumar-etal-2021-machine}
Sachin Kumar, Antonios Anastasopoulos, Shuly Wintner, and Yulia Tsvetkov.
  2021{\natexlab{a}}.
\newblock \href {https://doi.org/10.18653/v1/2021.acl-short.16} {Machine
  translation into low-resource language varieties}.
\newblock In \emph{Proceedings of the 59th Annual Meeting of the Association
  for Computational Linguistics and the 11th International Joint Conference on
  Natural Language Processing (Volume 2: Short Papers)}, pages 110--121,
  Online. Association for Computational Linguistics.

\bibitem[{Kumar et~al.(2021{\natexlab{b}})Kumar, Malmi, Severyn, and
  Tsvetkov}]{kumar2021controlled}
Sachin Kumar, Eric Malmi, Aliaksei Severyn, and Yulia Tsvetkov.
  2021{\natexlab{b}}.
\newblock Controlled text generation as continuous optimization with multiple
  constraints.
\newblock In \emph{Proc. NeurIPS}.

\bibitem[{Kumar et~al.(2022)Kumar, Paria, and Tsvetkov}]{kumar2022gradient}
Sachin Kumar, Biswajit Paria, and Yulia Tsvetkov. 2022.
\newblock Gradient-based constrained sampling from language models.
\newblock In \emph{Proceedings of the 2022 Conference on Empirical Methods in
  Natural Language Processing}, pages 2251--2277.

\bibitem[{Kumar et~al.(2019)Kumar, Wintner, Smith, and
  Tsvetkov}]{kumar-etal-2019-topics}
Sachin Kumar, Shuly Wintner, Noah~A. Smith, and Yulia Tsvetkov. 2019.
\newblock \href {https://doi.org/10.18653/v1/D19-1425} {Topics to avoid:
  Demoting latent confounds in text classification}.
\newblock In \emph{Proceedings of the 2019 Conference on Empirical Methods in
  Natural Language Processing and the 9th International Joint Conference on
  Natural Language Processing (EMNLP-IJCNLP)}, pages 4153--4163, Hong Kong,
  China. Association for Computational Linguistics.

\bibitem[{Kurita et~al.(2019)Kurita, Belova, and
  Anastasopoulos}]{kurita2019towards}
Keita Kurita, Anna Belova, and Antonios Anastasopoulos. 2019.
\newblock Towards robust toxic content classification.
\newblock \emph{arXiv preprint arXiv:1912.06872}.

\bibitem[{Lee et~al.(2022{\natexlab{a}})Lee, Park, Yoon, Bui, Dernoncourt, Kim,
  and Jung}]{lee2022factual}
Hwanhee Lee, Cheoneum Park, Seunghyun Yoon, Trung Bui, Franck Dernoncourt, Juae
  Kim, and Kyomin Jung. 2022{\natexlab{a}}.
\newblock \href {https://aclanthology.org/2022.gem-1.41} {Factual error
  correction for abstractive summaries using entity retrieval}.
\newblock In \emph{Proceedings of the 2nd Workshop on Natural Language
  Generation, Evaluation, and Metrics (GEM)}, pages 439--444, Abu Dhabi, United
  Arab Emirates (Hybrid). Association for Computational Linguistics.

\bibitem[{Lee et~al.(2022{\natexlab{b}})Lee, Ippolito, Nystrom, Zhang, Eck,
  Callison-Burch, and Carlini}]{lee-etal-2022-deduplicating}
Katherine Lee, Daphne Ippolito, Andrew Nystrom, Chiyuan Zhang, Douglas Eck,
  Chris Callison-Burch, and Nicholas Carlini. 2022{\natexlab{b}}.
\newblock \href {https://doi.org/10.18653/v1/2022.acl-long.577} {Deduplicating
  training data makes language models better}.
\newblock In \emph{Proceedings of the 60th Annual Meeting of the Association
  for Computational Linguistics (Volume 1: Long Papers)}, pages 8424--8445,
  Dublin, Ireland. Association for Computational Linguistics.

\bibitem[{Lehman et~al.(2021)Lehman, Jain, Pichotta, Goldberg, and
  Wallace}]{lehman-etal-2021-bert}
Eric Lehman, Sarthak Jain, Karl Pichotta, Yoav Goldberg, and Byron Wallace.
  2021.
\newblock \href {https://doi.org/10.18653/v1/2021.naacl-main.73} {Does {BERT}
  pretrained on clinical notes reveal sensitive data?}
\newblock In \emph{Proceedings of the 2021 Conference of the North American
  Chapter of the Association for Computational Linguistics: Human Language
  Technologies}, pages 946--959, Online. Association for Computational
  Linguistics.

\bibitem[{Leite et~al.(2020)Leite, Silva, Bontcheva, and
  Scarton}]{leite-etal-2020-toxic}
Jo{\~a}o~Augusto Leite, Diego Silva, Kalina Bontcheva, and Carolina Scarton.
  2020.
\newblock \href {https://aclanthology.org/2020.aacl-main.91} {Toxic language
  detection in social media for {B}razilian {P}ortuguese: New dataset and
  multilingual analysis}.
\newblock In \emph{Proceedings of the 1st Conference of the Asia-Pacific
  Chapter of the Association for Computational Linguistics and the 10th
  International Joint Conference on Natural Language Processing}, pages
  914--924, Suzhou, China. Association for Computational Linguistics.

\bibitem[{Levy et~al.(2021)Levy, Lazar, and
  Stanovsky}]{https://doi.org/10.48550/arxiv.2109.03858}
Shahar Levy, Koren Lazar, and Gabriel Stanovsky. 2021.
\newblock \href {https://doi.org/10.18653/v1/2021.findings-emnlp.211}
  {Collecting a large-scale gender bias dataset for coreference resolution and
  machine translation}.
\newblock In \emph{Findings of the Association for Computational Linguistics:
  EMNLP 2021}, pages 2470--2480, Punta Cana, Dominican Republic. Association
  for Computational Linguistics.

\bibitem[{Lewis et~al.(2020)Lewis, Perez, Piktus, Petroni, Karpukhin, Goyal,
  K\"{u}ttler, Lewis, Yih, Rockt\"{a}schel, Riedel, and
  Kiela}]{NEURIPS2020_6b493230}
Patrick Lewis, Ethan Perez, Aleksandra Piktus, Fabio Petroni, Vladimir
  Karpukhin, Naman Goyal, Heinrich K\"{u}ttler, Mike Lewis, Wen-tau Yih, Tim
  Rockt\"{a}schel, Sebastian Riedel, and Douwe Kiela. 2020.
\newblock \href
  {https://proceedings.neurips.cc/paper/2020/file/6b493230205f780e1bc26945df7481e5-Paper.pdf}
  {Retrieval-augmented generation for knowledge-intensive nlp tasks}.
\newblock In \emph{Advances in Neural Information Processing Systems},
  volume~33, pages 9459--9474. Curran Associates, Inc.

\bibitem[{Li et~al.(2018)Li, Zhu, Zhang, and Zong}]{li-etal-2018-ensure}
Haoran Li, Junnan Zhu, Jiajun Zhang, and Chengqing Zong. 2018.
\newblock \href {https://aclanthology.org/C18-1121} {Ensure the correctness of
  the summary: Incorporate entailment knowledge into abstractive sentence
  summarization}.
\newblock In \emph{Proceedings of the 27th International Conference on
  Computational Linguistics}, pages 1430--1441, Santa Fe, New Mexico, USA.
  Association for Computational Linguistics.

\bibitem[{Li et~al.(2022)Li, Tram{\`e}r, Liang, and Hashimoto}]{Li2021LargeLM}
Xuechen Li, Florian Tram{\`e}r, Percy Liang, and Tatsunori~B. Hashimoto. 2022.
\newblock Large language models can be strong differentially private learners.
\newblock \emph{ICLR}.

\bibitem[{Liang et~al.(2021)Liang, Wu, Morency, and
  Salakhutdinov}]{liang2021towards}
Paul~Pu Liang, Chiyu Wu, Louis-Philippe Morency, and Ruslan Salakhutdinov.
  2021.
\newblock Towards understanding and mitigating social biases in language
  models.
\newblock In \emph{International Conference on Machine Learning}, pages
  6565--6576. PMLR.

\bibitem[{Lin et~al.(2022{\natexlab{a}})Lin, Njoo, Field, Sharma, Reinecke,
  Althoff, and Tsvetkov}]{Lin2022GenderedMH}
Inna~Wanyin Lin, Lucille Njoo, Anjalie Field, Ashish Sharma, Katharina C.~H.
  Reinecke, Tim Althoff, and Yulia Tsvetkov. 2022{\natexlab{a}}.
\newblock Gendered mental health stigma in masked language models.
\newblock \emph{EMNLP}.

\bibitem[{Lin et~al.(2022{\natexlab{b}})Lin, Hilton, and
  Evans}]{lin2021truthfulqa}
Stephanie Lin, Jacob Hilton, and Owain Evans. 2022{\natexlab{b}}.
\newblock \href {https://doi.org/10.18653/v1/2022.acl-long.229}
  {{T}ruthful{QA}: Measuring how models mimic human falsehoods}.
\newblock In \emph{Proceedings of the 60th Annual Meeting of the Association
  for Computational Linguistics (Volume 1: Long Papers)}, pages 3214--3252,
  Dublin, Ireland. Association for Computational Linguistics.

\bibitem[{Lin et~al.(2021)Lin, Mihaylov, Artetxe, Wang, Chen, Simig, Ott,
  Goyal, Bhosale, Du, Pasunuru, Shleifer, Koura, Chaudhary, O'Horo, Wang,
  Zettlemoyer, Kozareva, Diab, Stoyanov, and Li}]{Lin2021FewshotLW}
Xi~Victoria Lin, Todor Mihaylov, Mikel Artetxe, Tianlu Wang, Shuohui Chen,
  Daniel Simig, Myle Ott, Naman Goyal, Shruti Bhosale, Jingfei Du, Ramakanth
  Pasunuru, Sam Shleifer, Punit~Singh Koura, Vishrav Chaudhary, Brian O'Horo,
  Jeff Wang, Luke Zettlemoyer, Zornitsa Kozareva, Mona Diab, Ves Stoyanov, and
  Xian Li. 2021.
\newblock Few-shot learning with multilingual language models.
\newblock \emph{ArXiv}, abs/2112.10668.

\bibitem[{Lipton(2018)}]{lipton2018mythos}
Zachary~C Lipton. 2018.
\newblock The mythos of model interpretability: In machine learning, the
  concept of interpretability is both important and slippery.
\newblock \emph{Queue}, 16(3):31--57.

\bibitem[{Liu et~al.(2021{\natexlab{a}})Liu, Sap, Lu, Swayamdipta, Bhagavatula,
  Smith, and Choi}]{liu-etal-2021-dexperts}
Alisa Liu, Maarten Sap, Ximing Lu, Swabha Swayamdipta, Chandra Bhagavatula,
  Noah~A. Smith, and Yejin Choi. 2021{\natexlab{a}}.
\newblock \href {https://doi.org/10.18653/v1/2021.acl-long.522} {{DE}xperts:
  Decoding-time controlled text generation with experts and anti-experts}.
\newblock In \emph{Proceedings of the 59th Annual Meeting of the Association
  for Computational Linguistics and the 11th International Joint Conference on
  Natural Language Processing (Volume 1: Long Papers)}, pages 6691--6706,
  Online. Association for Computational Linguistics.

\bibitem[{Liu et~al.(2020)Liu, Dacon, Fan, Liu, Liu, and
  Tang}]{liu-etal-2020-gender}
Haochen Liu, Jamell Dacon, Wenqi Fan, Hui Liu, Zitao Liu, and Jiliang Tang.
  2020.
\newblock \href {https://doi.org/10.18653/v1/2020.coling-main.390} {Does gender
  matter? towards fairness in dialogue systems}.
\newblock In \emph{Proceedings of the 28th International Conference on
  Computational Linguistics}, pages 4403--4416, Barcelona, Spain (Online).
  International Committee on Computational Linguistics.

\bibitem[{Liu et~al.(2023)Liu, Yuan, Fu, Jiang, Hayashi, and
  Neubig}]{https://doi.org/10.48550/arxiv.2107.13586}
Pengfei Liu, Weizhe Yuan, Jinlan Fu, Zhengbao Jiang, Hiroaki Hayashi, and
  Graham Neubig. 2023.
\newblock \href {https://doi.org/10.1145/3560815} {Pre-train, prompt, and
  predict: A systematic survey of prompting methods in natural language
  processing}.
\newblock \emph{ACM Comput. Surv.}, 55(9).

\bibitem[{Liu et~al.(2022)Liu, Yogatama, and Blunsom}]{10.1162/tacl_a_00476}
Qi~Liu, Dani Yogatama, and Phil Blunsom. 2022.
\newblock \href {https://doi.org/10.1162/tacl_a_00476} {{Relational
  Memory-Augmented Language Models}}.
\newblock \emph{Transactions of the Association for Computational Linguistics},
  10:555--572.

\bibitem[{Liu et~al.(2021{\natexlab{b}})Liu, Jia, Wei, Xu, Wang, and
  Vosoughi}]{Liu2021MitigatingPB}
Ruibo Liu, Chenyan Jia, Jason Wei, Guangxuan Xu, Lili Wang, and Soroush
  Vosoughi. 2021{\natexlab{b}}.
\newblock Mitigating political bias in language models through reinforced
  calibration.
\newblock In \emph{AAAI}.

\bibitem[{Liu and Forss(2015)}]{liu2015new}
Shuhua Liu and Thomas Forss. 2015.
\newblock New classification models for detecting hate and violence web
  content.
\newblock In \emph{2015 7th international joint conference on knowledge
  discovery, knowledge engineering and knowledge management (IC3K)}, volume~1,
  pages 487--495. IEEE.

\bibitem[{Lu et~al.(2022)Lu, Welleck, Hessel, Jiang, Qin, West, Ammanabrolu,
  and Choi}]{https://doi.org/10.48550/arxiv.2112.08726}
Ximing Lu, Sean Welleck, Jack Hessel, Liwei Jiang, Lianhui Qin, Peter West,
  Prithviraj Ammanabrolu, and Yejin Choi. 2022.
\newblock \href {https://openreview.net/forum?id=5HaIds3ux5O} {{QUARK}:
  Controllable text generation with reinforced unlearning}.
\newblock In \emph{Advances in Neural Information Processing Systems}.

\bibitem[{Lu et~al.(2021)Lu, West, Zellers, Le~Bras, Bhagavatula, and
  Choi}]{lu-etal-2021-neurologic}
Ximing Lu, Peter West, Rowan Zellers, Ronan Le~Bras, Chandra Bhagavatula, and
  Yejin Choi. 2021.
\newblock \href {https://doi.org/10.18653/v1/2021.naacl-main.339}
  {{N}euro{L}ogic decoding: (un)supervised neural text generation with
  predicate logic constraints}.
\newblock In \emph{Proceedings of the 2021 Conference of the North American
  Chapter of the Association for Computational Linguistics: Human Language
  Technologies}, pages 4288--4299, Online. Association for Computational
  Linguistics.

\bibitem[{Ma et~al.(2020)Ma, Sap, Rashkin, and
  Choi}]{ma-etal-2020-powertransformer}
Xinyao Ma, Maarten Sap, Hannah Rashkin, and Yejin Choi. 2020.
\newblock \href {https://doi.org/10.18653/v1/2020.emnlp-main.602}
  {{P}ower{T}ransformer: Unsupervised controllable revision for biased language
  correction}.
\newblock In \emph{Proceedings of the 2020 Conference on Empirical Methods in
  Natural Language Processing (EMNLP)}, pages 7426--7441, Online. Association
  for Computational Linguistics.

\bibitem[{Majmudar et~al.(2022)Majmudar, Dupuy, Peris, Smaili, Gupta, and
  Zemel}]{Majmudar2022DifferentiallyPD}
Jimit Majmudar, Christophe Dupuy, Charith~S. Peris, Sami Smaili, Rahul Gupta,
  and Richard~S. Zemel. 2022.
\newblock Differentially private decoding in large language models.
\newblock \emph{ArXiv}, abs/2205.13621.

\bibitem[{Mao et~al.(2020)Mao, Ren, Ji, and Han}]{Mao2020ConstrainedAS}
Yuning Mao, Xiang Ren, Heng Ji, and Jiawei Han. 2020.
\newblock Constrained abstractive summarization: Preserving factual consistency
  with constrained generation.
\newblock \emph{ArXiv}, abs/2010.12723.

\bibitem[{Martino et~al.(2020)Martino, Cresci, Barrón-Cedeño, Yu, Pietro, and
  Nakov}]{ijcai2020-0672}
Giovanni Da~San Martino, Stefano Cresci, Alberto Barrón-Cedeño, Seunghak Yu,
  Roberto~Di Pietro, and Preslav Nakov. 2020.
\newblock \href {https://doi.org/10.24963/ijcai.2020/672} {A survey on
  computational propaganda detection}.
\newblock In \emph{Proceedings of the Twenty-Ninth International Joint
  Conference on Artificial Intelligence, {IJCAI-20}}, pages 4826--4832.
  International Joint Conferences on Artificial Intelligence Organization.
\newblock Survey track.

\bibitem[{Mathew et~al.(2018)Mathew, Tharad, Rajgaria, Singhania, Maity, Goyal,
  and Mukherjee}]{https://doi.org/10.48550/arxiv.1808.04409}
Binny Mathew, Hardik Tharad, Subham Rajgaria, Prajwal Singhania, Suman~Kalyan
  Maity, Pawan Goyal, and Animesh Mukherjee. 2018.
\newblock Thou shalt not hate: Countering online hate speech.
\newblock In \emph{International Conference on Web and Social Media}.

\bibitem[{Maynez et~al.(2020)Maynez, Narayan, Bohnet, and
  McDonald}]{factentailment}
Joshua Maynez, Shashi Narayan, Bernd Bohnet, and Ryan McDonald. 2020.
\newblock \href {https://doi.org/10.18653/v1/2020.acl-main.173} {On
  faithfulness and factuality in abstractive summarization}.
\newblock In \emph{Proceedings of the 58th Annual Meeting of the Association
  for Computational Linguistics}, pages 1906--1919, Online. Association for
  Computational Linguistics.

\bibitem[{McCoy et~al.(2019)McCoy, Pavlick, and Linzen}]{McCoy2019RightFT}
R.~Thomas McCoy, Ellie Pavlick, and Tal Linzen. 2019.
\newblock Right for the wrong reasons: Diagnosing syntactic heuristics in
  natural language inference.
\newblock In \emph{Proc. ACL}.

\bibitem[{Meister et~al.(2022)Meister, Pimentel, Wiher, and
  Cotterell}]{typical}
Clara Meister, Tiago Pimentel, Gian Wiher, and Ryan Cotterell. 2022.
\newblock Typical decoding for natural language generation.
\newblock \emph{ArXiv}, abs/2202.00666.

\bibitem[{Meng et~al.(2022)Meng, Bau, Andonian, and
  Belinkov}]{meng2022locating}
Kevin Meng, David Bau, Alex~J Andonian, and Yonatan Belinkov. 2022.
\newblock \href {https://openreview.net/forum?id=-h6WAS6eE4} {Locating and
  editing factual associations in {GPT}}.
\newblock In \emph{Advances in Neural Information Processing Systems}.

\bibitem[{Meng et~al.(2023)Meng, Sharma, Andonian, Belinkov, and
  Bau}]{meng2023massediting}
Kevin Meng, Arnab~Sen Sharma, Alex~J Andonian, Yonatan Belinkov, and David Bau.
  2023.
\newblock \href {https://openreview.net/forum?id=MkbcAHIYgyS} {Mass-editing
  memory in a transformer}.
\newblock In \emph{International Conference on Learning Representations}.

\bibitem[{Mille et~al.(2021)Mille, Dhole, Mahamood, Perez-Beltrachini, Gangal,
  Kale, van Miltenburg, and Gehrmann}]{mille2021automatic}
Simon Mille, Kaustubh Dhole, Saad Mahamood, Laura Perez-Beltrachini, Varun
  Gangal, Mihir Kale, Emiel van Miltenburg, and Sebastian Gehrmann. 2021.
\newblock \href {https://openreview.net/forum?id=CSi1eu_2q96} {Automatic
  construction of evaluation suites for natural language generation datasets}.
\newblock In \emph{Thirty-fifth Conference on Neural Information Processing
  Systems Datasets and Benchmarks Track (Round 1)}.

\bibitem[{Mireshghallah et~al.(2022)Mireshghallah, Goyal, and
  Berg-Kirkpatrick}]{mireshghallah-etal-2022-mix}
Fatemehsadat Mireshghallah, Kartik Goyal, and Taylor Berg-Kirkpatrick. 2022.
\newblock \href {https://doi.org/10.18653/v1/2022.acl-long.31} {Mix and match:
  Learning-free controllable text generationusing energy language models}.
\newblock In \emph{Proceedings of the 60th Annual Meeting of the Association
  for Computational Linguistics (Volume 1: Long Papers)}, pages 401--415,
  Dublin, Ireland. Association for Computational Linguistics.

\bibitem[{Mirshghallah et~al.(2020)Mirshghallah, Taram, Vepakomma, Singh,
  Raskar, and Esmaeilzadeh}]{Mirshghallah2020PrivacyID}
Fatemehsadat Mirshghallah, Mohammadkazem Taram, Praneeth Vepakomma, Abhishek
  Singh, Ramesh Raskar, and Hadi Esmaeilzadeh. 2020.
\newblock Privacy in deep learning: A survey.
\newblock \emph{ArXiv}, abs/2004.12254.

\bibitem[{Mitchell et~al.(2023)Mitchell, Lee, Khazatsky, Manning, and
  Finn}]{mitchell2023detectgpt}
Eric Mitchell, Yoonho Lee, Alexander Khazatsky, Christopher~D Manning, and
  Chelsea Finn. 2023.
\newblock Detectgpt: Zero-shot machine-generated text detection using
  probability curvature.

\bibitem[{Mitchell et~al.(2022)Mitchell, Lin, Bosselut, Finn, and
  Manning}]{mitchell2022fast}
Eric Mitchell, Charles Lin, Antoine Bosselut, Chelsea Finn, and Christopher~D
  Manning. 2022.
\newblock \href {https://openreview.net/forum?id=0DcZxeWfOPt} {Fast model
  editing at scale}.
\newblock In \emph{International Conference on Learning Representations}.

\bibitem[{Moon et~al.(2020)Moon, Cho, and Lee}]{moon2020beep}
Jihyung Moon, Won~Ik Cho, and Junbum Lee. 2020.
\newblock \href {https://doi.org/10.18653/v1/2020.socialnlp-1.4} {{BEEP}!
  {K}orean corpus of online news comments for toxic speech detection}.
\newblock In \emph{Proceedings of the Eighth International Workshop on Natural
  Language Processing for Social Media}, pages 25--31, Online. Association for
  Computational Linguistics.

\bibitem[{Mubarak et~al.(2017)Mubarak, Darwish, and Magdy}]{mubarak2017abusive}
Hamdy Mubarak, Kareem Darwish, and Walid Magdy. 2017.
\newblock Abusive language detection on arabic social media.
\newblock In \emph{Proceedings of the first workshop on abusive language
  online}, pages 52--56.

\bibitem[{Nadeem et~al.(2021)Nadeem, Bethke, and
  Reddy}]{nadeem-etal-2021-stereoset}
Moin Nadeem, Anna Bethke, and Siva Reddy. 2021.
\newblock \href {https://doi.org/10.18653/v1/2021.acl-long.416} {{S}tereo{S}et:
  Measuring stereotypical bias in pretrained language models}.
\newblock In \emph{Proceedings of the 59th Annual Meeting of the Association
  for Computational Linguistics and the 11th International Joint Conference on
  Natural Language Processing (Volume 1: Long Papers)}, pages 5356--5371,
  Online. Association for Computational Linguistics.

\bibitem[{Nakov et~al.(2021)Nakov, Corney, Hasanain, Alam, Elsayed,
  Barr'on-Cedeno, Papotti, Shaar, and Martino}]{Nakov2021AutomatedFF}
Preslav Nakov, David Corney, Maram Hasanain, Firoj Alam, Tamer Elsayed, Alberto
  Barr'on-Cedeno, Paolo Papotti, Shaden Shaar, and Giovanni Da~San Martino.
  2021.
\newblock Automated fact-checking for assisting human fact-checkers.
\newblock In \emph{International Joint Conference on Artificial Intelligence}.

\bibitem[{Nan et~al.(2021)Nan, Nogueira~dos Santos, Zhu, Ng, McKeown,
  Nallapati, Zhang, Wang, Arnold, and Xiang}]{nan-etal-2021-improving}
Feng Nan, Cicero Nogueira~dos Santos, Henghui Zhu, Patrick Ng, Kathleen
  McKeown, Ramesh Nallapati, Dejiao Zhang, Zhiguo Wang, Andrew~O. Arnold, and
  Bing Xiang. 2021.
\newblock \href {https://doi.org/10.18653/v1/2021.acl-long.536} {Improving
  factual consistency of abstractive summarization via question answering}.
\newblock In \emph{Proceedings of the 59th Annual Meeting of the Association
  for Computational Linguistics and the 11th International Joint Conference on
  Natural Language Processing (Volume 1: Long Papers)}, pages 6881--6894,
  Online. Association for Computational Linguistics.

\bibitem[{Ngo et~al.(2021)Ngo, Raterink, Ara{\'u}jo, Zhang, Chen, Morisot, and
  Frosst}]{ngo2021mitigating}
Helen Ngo, Cooper Raterink, Jo{\~a}o~GM Ara{\'u}jo, Ivan Zhang, Carol Chen,
  Adrien Morisot, and Nicholas Frosst. 2021.
\newblock Mitigating harm in language models with conditional-likelihood
  filtration.
\newblock \emph{arXiv preprint arXiv:2108.07790}.

\bibitem[{Nobata et~al.(2016)Nobata, Tetreault, Thomas, Mehdad, and
  Chang}]{nobata2016abusive}
Chikashi Nobata, Joel Tetreault, Achint Thomas, Yashar Mehdad, and Yi~Chang.
  2016.
\newblock Abusive language detection in online user content.
\newblock In \emph{Proceedings of the 25th international conference on world
  wide web}, pages 145--153.

\bibitem[{Nozza et~al.(2021)Nozza, Bianchi, and Hovy}]{nozza-etal-2021-honest}
Debora Nozza, Federico Bianchi, and Dirk Hovy. 2021.
\newblock \href {https://doi.org/10.18653/v1/2021.naacl-main.191} {{HONEST}:
  Measuring hurtful sentence completion in language models}.
\newblock In \emph{Proceedings of the 2021 Conference of the North American
  Chapter of the Association for Computational Linguistics: Human Language
  Technologies}, pages 2398--2406, Online. Association for Computational
  Linguistics.

\bibitem[{Oshikawa et~al.(2020)Oshikawa, Qian, and
  Wang}]{oshikawa-etal-2020-survey}
Ray Oshikawa, Jing Qian, and William~Yang Wang. 2020.
\newblock \href {https://aclanthology.org/2020.lrec-1.747} {A survey on natural
  language processing for fake news detection}.
\newblock In \emph{Proceedings of the 12th Language Resources and Evaluation
  Conference}, pages 6086--6093, Marseille, France. European Language Resources
  Association.

\bibitem[{Ousidhoum et~al.(2019{\natexlab{a}})Ousidhoum, Lin, Zhang, Song, and
  Yeung}]{ousidhoum2019multilingual}
Nedjma Ousidhoum, Zizheng Lin, Hongming Zhang, Yangqiu Song, and Dit-Yan Yeung.
  2019{\natexlab{a}}.
\newblock \href {https://doi.org/10.18653/v1/D19-1474} {Multilingual and
  multi-aspect hate speech analysis}.
\newblock In \emph{Proceedings of the 2019 Conference on Empirical Methods in
  Natural Language Processing and the 9th International Joint Conference on
  Natural Language Processing (EMNLP-IJCNLP)}, pages 4675--4684, Hong Kong,
  China. Association for Computational Linguistics.

\bibitem[{Ousidhoum et~al.(2019{\natexlab{b}})Ousidhoum, Lin, Zhang, Song, and
  Yeung}]{ousidhoum-etal-2019-multilingual}
Nedjma Ousidhoum, Zizheng Lin, Hongming Zhang, Yangqiu Song, and Dit-Yan Yeung.
  2019{\natexlab{b}}.
\newblock \href {https://doi.org/10.18653/v1/D19-1474} {Multilingual and
  multi-aspect hate speech analysis}.
\newblock In \emph{Proceedings of the 2019 Conference on Empirical Methods in
  Natural Language Processing and the 9th International Joint Conference on
  Natural Language Processing (EMNLP-IJCNLP)}, pages 4675--4684, Hong Kong,
  China. Association for Computational Linguistics.

\bibitem[{Ouyang et~al.(2022)Ouyang, Wu, Jiang, Almeida, Wainwright, Mishkin,
  Zhang, Agarwal, Slama, Ray, Schulman, Hilton, Kelton, Miller, Simens, Askell,
  Welinder, Christiano, Leike, and Lowe}]{Ouyang2022TrainingLM}
Long Ouyang, Jeff Wu, Xu~Jiang, Diogo Almeida, Carroll~L. Wainwright, Pamela
  Mishkin, Chong Zhang, Sandhini Agarwal, Katarina Slama, Alex Ray, John
  Schulman, Jacob Hilton, Fraser Kelton, Luke~E. Miller, Maddie Simens, Amanda
  Askell, Peter Welinder, Paul~Francis Christiano, Jan Leike, and Ryan~J. Lowe.
  2022.
\newblock Training language models to follow instructions with human feedback.
\newblock \emph{NeurIPS}.

\bibitem[{Pagnoni et~al.(2021)Pagnoni, Balachandran, and Tsvetkov}]{frank}
Artidoro Pagnoni, Vidhisha Balachandran, and Yulia Tsvetkov. 2021.
\newblock \href {https://doi.org/10.18653/v1/2021.naacl-main.383}
  {Understanding factuality in abstractive summarization with {FRANK}: A
  benchmark for factuality metrics}.
\newblock In \emph{Proceedings of the 2021 Conference of the North American
  Chapter of the Association for Computational Linguistics: Human Language
  Technologies}, pages 4812--4829, Online. Association for Computational
  Linguistics.

\bibitem[{Pamungkas et~al.(2021)Pamungkas, Basile, and
  Patti}]{pamungkas2021towards}
Endang~Wahyu Pamungkas, Valerio Basile, and Viviana Patti. 2021.
\newblock Towards multidomain and multilingual abusive language detection: a
  survey.
\newblock \emph{Personal and Ubiquitous Computing}, pages 1--27.

\bibitem[{Pascual et~al.(2021)Pascual, Egressy, Meister, Cotterell, and
  Wattenhofer}]{pascual-etal-2021-plug-play}
Damian Pascual, Beni Egressy, Clara Meister, Ryan Cotterell, and Roger
  Wattenhofer. 2021.
\newblock \href {https://doi.org/10.18653/v1/2021.findings-emnlp.334} {A
  plug-and-play method for controlled text generation}.
\newblock In \emph{Findings of the Association for Computational Linguistics:
  EMNLP 2021}, pages 3973--3997, Punta Cana, Dominican Republic. Association
  for Computational Linguistics.

\bibitem[{Patel and Lam(2023)}]{Patel2023ChatGPTTF}
Sajan Patel and Kyle Lam. 2023.
\newblock Chatgpt: the future of discharge summaries?
\newblock \emph{The Lancet. Digital health}.

\bibitem[{Pavlopoulos et~al.(2017)Pavlopoulos, Malakasiotis, and
  Androutsopoulos}]{pavlopoulos2017deep}
John Pavlopoulos, Prodromos Malakasiotis, and Ion Androutsopoulos. 2017.
\newblock \href {https://doi.org/10.18653/v1/W17-3004} {Deep learning for user
  comment moderation}.
\newblock In \emph{Proceedings of the First Workshop on Abusive Language
  Online}, pages 25--35, Vancouver, BC, Canada. Association for Computational
  Linguistics.

\bibitem[{Perez et~al.(2022)Perez, Huang, Song, Cai, Ring, Aslanides, Glaese,
  McAleese, and Irving}]{Perez2022RedTL}
Ethan Perez, Saffron Huang, Francis Song, Trevor Cai, Roman Ring, John
  Aslanides, Amelia Glaese, Nathan McAleese, and Geoffrey Irving. 2022.
\newblock Red teaming language models with language models.
\newblock In \emph{Conference on Empirical Methods in Natural Language
  Processing}.

\bibitem[{Peters et~al.(2018)Peters, Neumann, Iyyer, Gardner, Clark, Lee, and
  Zettlemoyer}]{peters-etal-2018-deep}
Matthew~E. Peters, Mark Neumann, Mohit Iyyer, Matt Gardner, Christopher Clark,
  Kenton Lee, and Luke Zettlemoyer. 2018.
\newblock \href {https://doi.org/10.18653/v1/N18-1202} {Deep contextualized
  word representations}.
\newblock In \emph{Proceedings of the 2018 Conference of the North {A}merican
  Chapter of the Association for Computational Linguistics: Human Language
  Technologies, Volume 1 (Long Papers)}, pages 2227--2237, New Orleans,
  Louisiana. Association for Computational Linguistics.

\bibitem[{Pitsilis et~al.(2018)Pitsilis, Ramampiaro, and
  Langseth}]{pitsilis2018effective}
Georgios~K Pitsilis, Heri Ramampiaro, and Helge Langseth. 2018.
\newblock Effective hate-speech detection in twitter data using recurrent
  neural networks.
\newblock \emph{Applied Intelligence}, 48(12):4730--4742.

\bibitem[{Prabhumoye et~al.(2018)Prabhumoye, Tsvetkov, Salakhutdinov, and
  Black}]{prabhumoye2018style}
Shrimai Prabhumoye, Yulia Tsvetkov, Ruslan Salakhutdinov, and Alan~W Black.
  2018.
\newblock \href {https://doi.org/10.18653/v1/P18-1080} {Style transfer through
  back-translation}.
\newblock In \emph{Proceedings of the 56th Annual Meeting of the Association
  for Computational Linguistics (Volume 1: Long Papers)}, pages 866--876,
  Melbourne, Australia. Association for Computational Linguistics.

\bibitem[{Pryzant et~al.(2020)Pryzant, Diehl~Martinez, Dass, Kurohashi,
  Jurafsky, and Yang}]{Pryzant_DiehlMartinez_Dass_Kurohashi_Jurafsky_Yang_2020}
Reid Pryzant, Richard Diehl~Martinez, Nathan Dass, Sadao Kurohashi, Dan
  Jurafsky, and Diyi Yang. 2020.
\newblock \href {https://doi.org/10.1609/aaai.v34i01.5385} {Automatically
  neutralizing subjective bias in text}.
\newblock \emph{Proceedings of the AAAI Conference on Artificial Intelligence},
  34(01):480--489.

\bibitem[{Qin et~al.(2020)Qin, Shwartz, West, Bhagavatula, Hwang, Le~Bras,
  Bosselut, and Choi}]{qin-etal-2020-back}
Lianhui Qin, Vered Shwartz, Peter West, Chandra Bhagavatula, Jena~D. Hwang,
  Ronan Le~Bras, Antoine Bosselut, and Yejin Choi. 2020.
\newblock \href {https://doi.org/10.18653/v1/2020.emnlp-main.58} {Back to the
  future: Unsupervised backprop-based decoding for counterfactual and abductive
  commonsense reasoning}.
\newblock In \emph{Proceedings of the 2020 Conference on Empirical Methods in
  Natural Language Processing (EMNLP)}, pages 794--805, Online. Association for
  Computational Linguistics.

\bibitem[{Rae et~al.(2021)Rae, Borgeaud, Cai, Millican, Hoffmann, Song,
  Aslanides, Henderson, Ring, Young, Rutherford, Hennigan, Menick, Cassirer,
  Powell, van~den Driessche, Hendricks, Rauh, Huang, Glaese, Welbl, Dathathri,
  Huang, Uesato, Mellor, Higgins, Creswell, McAleese, Wu, Elsen, Jayakumar,
  Buchatskaya, Budden, Sutherland, Simonyan, Paganini, Sifre, Martens, Li,
  Kuncoro, Nematzadeh, Gribovskaya, Donato, Lazaridou, Mensch, Lespiau,
  Tsimpoukelli, Grigorev, Fritz, Sottiaux, Pajarskas, Pohlen, Gong, Toyama,
  de~Masson~d'Autume, Li, Terzi, Mikulik, Babuschkin, Clark, de~Las~Casas, Guy,
  Jones, Bradbury, Johnson, Hechtman, Weidinger, Gabriel, Isaac, Lockhart,
  Osindero, Rimell, Dyer, Vinyals, Ayoub, Stanway, Bennett, Hassabis,
  Kavukcuoglu, and Irving}]{DBLP:journals/corr/abs-2112-11446}
Jack~W. Rae, Sebastian Borgeaud, Trevor Cai, Katie Millican, Jordan Hoffmann,
  H.~Francis Song, John Aslanides, Sarah Henderson, Roman Ring, Susannah Young,
  Eliza Rutherford, Tom Hennigan, Jacob Menick, Albin Cassirer, Richard Powell,
  George van~den Driessche, Lisa~Anne Hendricks, Maribeth Rauh, Po{-}Sen Huang,
  Amelia Glaese, Johannes Welbl, Sumanth Dathathri, Saffron Huang, Jonathan
  Uesato, John Mellor, Irina Higgins, Antonia Creswell, Nat McAleese, Amy Wu,
  Erich Elsen, Siddhant~M. Jayakumar, Elena Buchatskaya, David Budden, Esme
  Sutherland, Karen Simonyan, Michela Paganini, Laurent Sifre, Lena Martens,
  Xiang~Lorraine Li, Adhiguna Kuncoro, Aida Nematzadeh, Elena Gribovskaya,
  Domenic Donato, Angeliki Lazaridou, Arthur Mensch, Jean{-}Baptiste Lespiau,
  Maria Tsimpoukelli, Nikolai Grigorev, Doug Fritz, Thibault Sottiaux, Mantas
  Pajarskas, Toby Pohlen, Zhitao Gong, Daniel Toyama, Cyprien
  de~Masson~d'Autume, Yujia Li, Tayfun Terzi, Vladimir Mikulik, Igor
  Babuschkin, Aidan Clark, Diego de~Las~Casas, Aurelia Guy, Chris Jones, James
  Bradbury, Matthew Johnson, Blake~A. Hechtman, Laura Weidinger, Iason Gabriel,
  William~S. Isaac, Edward Lockhart, Simon Osindero, Laura Rimell, Chris Dyer,
  Oriol Vinyals, Kareem Ayoub, Jeff Stanway, Lorrayne Bennett, Demis Hassabis,
  Koray Kavukcuoglu, and Geoffrey Irving. 2021.
\newblock \href {https://arxiv.org/abs/2112.11446} {Scaling language models:
  Methods, analysis {\&} insights from training gopher}.
\newblock arXiv.

\bibitem[{Raffel et~al.(2020)Raffel, Shazeer, Roberts, Lee, Narang, Matena,
  Zhou, Li, and Liu}]{JMLR:v21:20-074}
Colin Raffel, Noam Shazeer, Adam Roberts, Katherine Lee, Sharan Narang, Michael
  Matena, Yanqi Zhou, Wei Li, and Peter~J. Liu. 2020.
\newblock \href {http://jmlr.org/papers/v21/20-074.html} {Exploring the limits
  of transfer learning with a unified text-to-text transformer}.
\newblock \emph{Journal of Machine Learning Research}, 21(140):1--67.

\bibitem[{Raji et~al.(2020)Raji, Smart, White, Mitchell, Gebru, Hutchinson,
  Smith-Loud, Theron, and Barnes}]{raji2020closing}
Inioluwa~Deborah Raji, Andrew Smart, Rebecca~N White, Margaret Mitchell, Timnit
  Gebru, Ben Hutchinson, Jamila Smith-Loud, Daniel Theron, and Parker Barnes.
  2020.
\newblock Closing the ai accountability gap: Defining an end-to-end framework
  for internal algorithmic auditing.
\newblock In \emph{Proceedings of the 2020 conference on fairness,
  accountability, and transparency}, pages 33--44.

\bibitem[{Ramamurthy et~al.(2022)Ramamurthy, Ammanabrolu, Brantley, Hessel,
  Sifa, Bauckhage, Hajishirzi, and
  Choi}]{https://doi.org/10.48550/arxiv.2210.01241}
Rajkumar Ramamurthy, Prithviraj Ammanabrolu, Kianté Brantley, Jack Hessel,
  Rafet Sifa, Christian Bauckhage, Hannaneh Hajishirzi, and Yejin Choi. 2022.
\newblock \href {https://doi.org/10.48550/ARXIV.2210.01241} {Is reinforcement
  learning (not) for natural language processing?: Benchmarks, baselines, and
  building blocks for natural language policy optimization}.
\newblock arXiv.

\bibitem[{Rauh et~al.(2022)Rauh, Mellor, Uesato, Huang, Welbl, Weidinger,
  Dathathri, Glaese, Irving, Gabriel, Isaac, and
  Hendricks}]{rauh2022characteristics}
Maribeth Rauh, John F~J Mellor, Jonathan Uesato, Po-Sen Huang, Johannes Welbl,
  Laura Weidinger, Sumanth Dathathri, Amelia Glaese, Geoffrey Irving, Iason
  Gabriel, William Isaac, and Lisa~Anne Hendricks. 2022.
\newblock \href {https://openreview.net/forum?id=u46CbCaLufp} {Characteristics
  of harmful text: Towards rigorous benchmarking of language models}.
\newblock In \emph{Thirty-sixth Conference on Neural Information Processing
  Systems Datasets and Benchmarks Track}.

\bibitem[{Reuters(2023)}]{chatgpt-india}
Reuters. 2023.
\newblock \href
  {https://indianexpress.com/article/technology/chatgpt-sets-record-for-fastest-growing-user-base-8419389/}
  {Chatgpt sets record for fastest-growing user base}.

\bibitem[{Ribeiro et~al.(2020)Ribeiro, Wu, Guestrin, and
  Singh}]{ribeiro-etal-2020-beyond}
Marco~Tulio Ribeiro, Tongshuang Wu, Carlos Guestrin, and Sameer Singh. 2020.
\newblock \href {https://doi.org/10.18653/v1/2020.acl-main.442} {Beyond
  accuracy: Behavioral testing of {NLP} models with {C}heck{L}ist}.
\newblock In \emph{Proceedings of the 58th Annual Meeting of the Association
  for Computational Linguistics}, pages 4902--4912, Online. Association for
  Computational Linguistics.

\bibitem[{Ross et~al.(2017)Ross, Rist, Carbonell, Cabrera, Kurowsky, and
  Wojatzki}]{ross2017measuring}
Bj{\"o}rn Ross, Michael Rist, Guillermo Carbonell, Benjamin Cabrera, Nils
  Kurowsky, and Michael Wojatzki. 2017.
\newblock Measuring the reliability of hate speech annotations: The case of the
  european refugee crisis.
\newblock \emph{arXiv preprint arXiv:1701.08118}.

\bibitem[{Sabri et~al.(2021)Sabri, Basile, Caselli
  et~al.}]{gnrem2021LeveragingBI}
Nazanin Sabri, Valerio Basile, Tommaso Caselli, et~al. 2021.
\newblock Leveraging bias in pre-trained word embeddings for unsupervised
  microaggression detection.
\newblock In \emph{Italian Conference on Computational Linguistics 2021:
  CLiC-it 2021}. CEUR Workshop Proceedings (CEUR-WS. org).

\bibitem[{Sanh et~al.(2022)Sanh, Webson, Raffel, Bach, Sutawika, Alyafeai,
  Chaffin, Stiegler, Raja, Dey, Bari, Xu, Thakker, Sharma, Szczechla, Kim,
  Chhablani, Nayak, Datta, Chang, Jiang, Wang, Manica, Shen, Yong, Pandey,
  Bawden, Wang, Neeraj, Rozen, Sharma, Santilli, Fevry, Fries, Teehan, Scao,
  Biderman, Gao, Wolf, and Rush}]{sanh2022multitask}
Victor Sanh, Albert Webson, Colin Raffel, Stephen Bach, Lintang Sutawika, Zaid
  Alyafeai, Antoine Chaffin, Arnaud Stiegler, Arun Raja, Manan Dey, M~Saiful
  Bari, Canwen Xu, Urmish Thakker, Shanya~Sharma Sharma, Eliza Szczechla,
  Taewoon Kim, Gunjan Chhablani, Nihal Nayak, Debajyoti Datta, Jonathan Chang,
  Mike Tian-Jian Jiang, Han Wang, Matteo Manica, Sheng Shen, Zheng~Xin Yong,
  Harshit Pandey, Rachel Bawden, Thomas Wang, Trishala Neeraj, Jos Rozen,
  Abheesht Sharma, Andrea Santilli, Thibault Fevry, Jason~Alan Fries, Ryan
  Teehan, Teven~Le Scao, Stella Biderman, Leo Gao, Thomas Wolf, and Alexander~M
  Rush. 2022.
\newblock \href {https://openreview.net/forum?id=9Vrb9D0WI4} {Multitask
  prompted training enables zero-shot task generalization}.
\newblock In \emph{International Conference on Learning Representations}.

\bibitem[{Sap et~al.(2019)Sap, Card, Gabriel, Choi, and Smith}]{sap2019risk}
Maarten Sap, Dallas Card, Saadia Gabriel, Yejin Choi, and A~Noah Smith. 2019.
\newblock The risk of racial bias in hate speech detection.
\newblock In \emph{ACL}.

\bibitem[{Sap et~al.(2021)Sap, Swayamdipta, Vianna, Zhou, Choi, and
  Smith}]{sap2021annotators}
Maarten Sap, Swabha Swayamdipta, Laura Vianna, Xuhui Zhou, Yejin Choi, and
  Noah~A Smith. 2021.
\newblock Annotators with attitudes: How annotator beliefs and identities bias
  toxic language detection.
\newblock \emph{arXiv preprint arXiv:2111.07997}.

\bibitem[{Schick et~al.(2021)Schick, Udupa, and Schütze}]{schick2020self}
Timo Schick, Sahana Udupa, and Hinrich Schütze. 2021.
\newblock \href {http://arxiv.org/abs/2103.00453} {Self-diagnosis and
  self-debiasing: A proposal for reducing corpus-based bias in nlp}.
\newblock \emph{Computing Research Repository}, arXiv:2103.00453.

\bibitem[{Schulten(2023)}]{chatgpt-education}
Katherine Schulten. 2023.
\newblock \href
  {https://www.nytimes.com/2023/01/24/learning/how-should-schools-respond-to-chatgpt.html}
  {How should schools respond to chatgpt?}

\bibitem[{Scialom et~al.(2021)Scialom, Dray, Gallinari, Lamprier, Piwowarski,
  Staiano, and Wang}]{Scialom2021QuestEvalSA}
Thomas Scialom, Paul-Alexis Dray, Patrick Gallinari, Sylvain Lamprier, Benjamin
  Piwowarski, Jacopo Staiano, and Alex Wang. 2021.
\newblock Questeval: Summarization asks for fact-based evaluation.
\newblock In \emph{Conference on Empirical Methods in Natural Language
  Processing}.

\bibitem[{Shaar et~al.(2021)Shaar, Alam, Martino, and
  Nakov}]{Shaar2021AssistingTH}
Shaden Shaar, Firoj Alam, Giovanni Da~San Martino, and Preslav Nakov. 2021.
\newblock Assisting the human fact-checkers: Detecting all previously
  fact-checked claims in a document.
\newblock In \emph{Conference on Empirical Methods in Natural Language
  Processing}.

\bibitem[{Shao et~al.(2018)Shao, Ciampaglia, Varol, Yang, Flammini, and
  Menczer}]{shao2018spread}
Chengcheng Shao, Giovanni~Luca Ciampaglia, Onur Varol, Kai-Cheng Yang,
  Alessandro Flammini, and Filippo Menczer. 2018.
\newblock The spread of low-credibility content by social bots.
\newblock \emph{Nature communications}, 9(1):1--9.

\bibitem[{Sheng et~al.(2021)Sheng, Chang, Natarajan, and
  Peng}]{sheng-etal-2021-societal}
Emily Sheng, Kai-Wei Chang, Prem Natarajan, and Nanyun Peng. 2021.
\newblock \href {https://doi.org/10.18653/v1/2021.acl-long.330} {Societal
  biases in language generation: Progress and challenges}.
\newblock In \emph{Proceedings of the 59th Annual Meeting of the Association
  for Computational Linguistics and the 11th International Joint Conference on
  Natural Language Processing (Volume 1: Long Papers)}, pages 4275--4293,
  Online. Association for Computational Linguistics.

\bibitem[{Shi et~al.(2021)Shi, Cui, Li, Jia, and Yu}]{Shi2021SelectiveDP}
Weiyan Shi, Aiqi Cui, Evan Li, R.~Jia, and Zhou Yu. 2021.
\newblock Selective differential privacy for language modeling.
\newblock In \emph{North American Chapter of the Association for Computational
  Linguistics}.

\bibitem[{Shliazhko et~al.(2022)Shliazhko, Fenogenova, Tikhonova, Mikhailov,
  Kozlova, and Shavrina}]{https://doi.org/10.48550/arxiv.2204.07580}
Oleh Shliazhko, Alena Fenogenova, Maria Tikhonova, Vladislav Mikhailov,
  Anastasia Kozlova, and Tatiana Shavrina. 2022.
\newblock \href {https://doi.org/10.48550/ARXIV.2204.07580} {mgpt: Few-shot
  learners go multilingual}.
\newblock arXiv.

\bibitem[{Shoeybi et~al.(2019)Shoeybi, Patwary, Puri, LeGresley, Casper, and
  Catanzaro}]{Shoeybi2019MegatronLMTM}
Mohammad Shoeybi, Mostofa Patwary, Raul Puri, Patrick LeGresley, Jared Casper,
  and Bryan Catanzaro. 2019.
\newblock Megatron-lm: Training multi-billion parameter language models using
  model parallelism.
\newblock \emph{ArXiv}, abs/1909.08053.

\bibitem[{Simard et~al.(2007)Simard, Ueffing, Isabelle, and
  Kuhn}]{Simard2007RuleBasedTW}
Michel Simard, Nicola Ueffing, Pierre Isabelle, and Roland Kuhn. 2007.
\newblock Rule-based translation with statistical phrase-based post-editing.
\newblock In \emph{WMT@ACL}.

\bibitem[{So-hyun(2023)}]{chatgpt-korea}
Kim So-hyun. 2023.
\newblock \href {hhttps://www.koreaherald.com/view.php?ud=20230201000790} {[the
  korean dilemma] what chatgpt means for korea}.

\bibitem[{Solaiman et~al.(2019)Solaiman, Brundage, Clark, Askell, Herbert-Voss,
  Wu, Radford, Krueger, Kim, Kreps et~al.}]{solaiman2019release}
Irene Solaiman, Miles Brundage, Jack Clark, Amanda Askell, Ariel Herbert-Voss,
  Jeff Wu, Alec Radford, Gretchen Krueger, Jong~Wook Kim, Sarah Kreps, et~al.
  2019.
\newblock Release strategies and the social impacts of language models.
\newblock \emph{arXiv preprint arXiv:1908.09203}.

\bibitem[{Stafanovi{\v{c}}s et~al.(2020)Stafanovi{\v{c}}s, Bergmanis, and
  Pinnis}]{stafanovics-etal-2020-mitigating}
Art{\=u}rs Stafanovi{\v{c}}s, Toms Bergmanis, and M{\=a}rcis Pinnis. 2020.
\newblock \href {https://aclanthology.org/2020.wmt-1.73} {Mitigating gender
  bias in machine translation with target gender annotations}.
\newblock In \emph{Proceedings of the Fifth Conference on Machine Translation},
  pages 629--638, Online. Association for Computational Linguistics.

\bibitem[{Stiennon et~al.(2020)Stiennon, Ouyang, Wu, Ziegler, Lowe, Voss,
  Radford, Amodei, and Christiano}]{Stiennon2020LearningTS}
Nisan Stiennon, Long Ouyang, Jeff Wu, Daniel~M. Ziegler, Ryan~J. Lowe, Chelsea
  Voss, Alec Radford, Dario Amodei, and Paul Christiano. 2020.
\newblock Learning to summarize from human feedback.
\newblock \emph{NeurIPS}, abs/2009.01325.

\bibitem[{Stokel-Walker(2023)}]{chatgpt-science}
Chris Stokel-Walker. 2023.
\newblock \href {https://www.nature.com/articles/d41586-023-00107-z} {Chatgpt
  listed as author on research papers: many scientists disapprove}.

\bibitem[{Sun et~al.(2020)Sun, Ho, and yi~Lee}]{Sun2020LAMOLLM}
Fan-Keng Sun, Cheng-Hao Ho, and Hung yi~Lee. 2020.
\newblock Lamol: Language modeling for lifelong language learning.
\newblock In \emph{ICLR}.

\bibitem[{Sun et~al.(2019)Sun, Gaut, Tang, Huang, ElSherief, Zhao, Mirza,
  Belding, Chang, and Wang}]{sun-etal-2019-mitigating}
Tony Sun, Andrew Gaut, Shirlyn Tang, Yuxin Huang, Mai ElSherief, Jieyu Zhao,
  Diba Mirza, Elizabeth Belding, Kai-Wei Chang, and William~Yang Wang. 2019.
\newblock \href {https://doi.org/10.18653/v1/P19-1159} {Mitigating gender bias
  in natural language processing: Literature review}.
\newblock In \emph{Proceedings of the 57th Annual Meeting of the Association
  for Computational Linguistics}, pages 1630--1640, Florence, Italy.
  Association for Computational Linguistics.

\bibitem[{Taori and Hashimoto(2022)}]{taori2022data}
Rohan Taori and Tatsunori Hashimoto. 2022.
\newblock \href {https://openreview.net/forum?id=OJCNhcBFfp} {Data feedback
  loops: Model-driven amplification of dataset biases}.
\newblock In \emph{NeurIPS 2022 Workshop on Distribution Shifts: Connecting
  Methods and Applications}.

\bibitem[{Thorne et~al.(2018)Thorne, Vlachos, Christodoulopoulos, and
  Mittal}]{thorne-etal-2018-fever}
James Thorne, Andreas Vlachos, Christos Christodoulopoulos, and Arpit Mittal.
  2018.
\newblock \href {https://doi.org/10.18653/v1/N18-1074} {{FEVER}: a large-scale
  dataset for fact extraction and {VER}ification}.
\newblock In \emph{Proceedings of the 2018 Conference of the North {A}merican
  Chapter of the Association for Computational Linguistics: Human Language
  Technologies, Volume 1 (Long Papers)}, pages 809--819, New Orleans,
  Louisiana. Association for Computational Linguistics.

\bibitem[{Varghese(2023)}]{chatgpt-india2}
Ranjana~Mary Varghese. 2023.
\newblock \href
  {https://www.deccanherald.com/opinion/will-chatgpt-shake-up-higher-education-in-india-1187775.html}
  {Will chatgpt shake up higher education in india?}

\bibitem[{Vaswani et~al.(2017)Vaswani, Shazeer, Parmar, Uszkoreit, Jones,
  Gomez, Kaiser, and Polosukhin}]{NIPS2017_3f5ee243}
Ashish Vaswani, Noam Shazeer, Niki Parmar, Jakob Uszkoreit, Llion Jones,
  Aidan~N Gomez, \L~ukasz Kaiser, and Illia Polosukhin. 2017.
\newblock \href
  {https://proceedings.neurips.cc/paper/2017/file/3f5ee243547dee91fbd053c1c4a845aa-Paper.pdf}
  {Attention is all you need}.
\newblock In \emph{Advances in Neural Information Processing Systems},
  volume~30. Curran Associates, Inc.

\bibitem[{Vincent(2022)}]{gpt-4chan}
James Vincent. 2022.
\newblock \href
  {https://www.theverge.com/2022/6/8/23159465/youtuber-ai-bot-pol-gpt-4chan-yannic-kilcher-ethics}
  {{YouTuber trains AI bot on 4chan's pile o' bile with entirely predictable
  results}}.

\bibitem[{Wan and Bansal(2022)}]{wan2022factpegasus}
David Wan and Mohit Bansal. 2022.
\newblock Factpegasus: Factuality-aware pre-training and fine-tuning for
  abstractive summarization.
\newblock In \emph{NAACL 2022}.

\bibitem[{Wang et~al.(2020)Wang, Cho, and Lewis}]{factqa}
Alex Wang, Kyunghyun Cho, and Mike Lewis. 2020.
\newblock \href {https://doi.org/10.18653/v1/2020.acl-main.450} {Asking and
  answering questions to evaluate the factual consistency of summaries}.
\newblock In \emph{Proceedings of the 58th Annual Meeting of the Association
  for Computational Linguistics}, pages 5008--5020, Online. Association for
  Computational Linguistics.

\bibitem[{Wang et~al.(2019{\natexlab{a}})Wang, Pruksachatkun, Nangia, Singh,
  Michael, Hill, Levy, and Bowman}]{superglue}
Alex Wang, Yada Pruksachatkun, Nikita Nangia, Amanpreet Singh, Julian Michael,
  Felix Hill, Omer Levy, and Samuel Bowman. 2019{\natexlab{a}}.
\newblock \href
  {https://proceedings.neurips.cc/paper/2019/file/4496bf24afe7fab6f046bf4923da8de6-Paper.pdf}
  {Superglue: A stickier benchmark for general-purpose language understanding
  systems}.
\newblock In \emph{Advances in Neural Information Processing Systems},
  volume~32. Curran Associates, Inc.

\bibitem[{Wang et~al.(2019{\natexlab{b}})Wang, Singh, Michael, Hill, Levy, and
  Bowman}]{wang2018glue}
Alex Wang, Amanpreet Singh, Julian Michael, Felix Hill, Omer Levy, and
  Samuel~R. Bowman. 2019{\natexlab{b}}.
\newblock \href {https://openreview.net/forum?id=rJ4km2R5t7} {{GLUE}: A
  multi-task benchmark and analysis platform for natural language
  understanding}.
\newblock In \emph{International Conference on Learning Representations}.

\bibitem[{Wang et~al.(2022)Wang, Ping, Xiao, Xu, Patwary, Shoeybi, Li,
  Anandkumar, and Catanzaro}]{https://doi.org/10.48550/arxiv.2202.04173}
Boxin Wang, Wei Ping, Chaowei Xiao, Peng Xu, Mostofa Patwary, Mohammad Shoeybi,
  Bo~Li, Anima Anandkumar, and Bryan Catanzaro. 2022.
\newblock Exploring the limits of domain-adaptive training for detoxifying
  large-scale language models.
\newblock \emph{NeurIPS}.

\bibitem[{Wang et~al.(2021{\natexlab{a}})Wang, Liu, Zhu, Shou, Gong, Xu, and
  Zeng}]{wang-etal-2021-retrieval-enhanced}
Han Wang, Yang Liu, Chenguang Zhu, Linjun Shou, Ming Gong, Yichong Xu, and
  Michael Zeng. 2021{\natexlab{a}}.
\newblock \href {https://doi.org/10.18653/v1/2021.findings-acl.269} {Retrieval
  enhanced model for commonsense generation}.
\newblock In \emph{Findings of the Association for Computational Linguistics:
  ACL-IJCNLP 2021}, pages 3056--3062, Online. Association for Computational
  Linguistics.

\bibitem[{Wang et~al.(2021{\natexlab{b}})Wang, Li, Aslan, and
  Vinyals}]{wang-etal-2021-wikigraphs}
Luyu Wang, Yujia Li, Ozlem Aslan, and Oriol Vinyals. 2021{\natexlab{b}}.
\newblock \href {https://doi.org/10.18653/v1/2021.textgraphs-1.7}
  {{W}iki{G}raphs: A {W}ikipedia text - knowledge graph paired dataset}.
\newblock In \emph{Proceedings of the Fifteenth Workshop on Graph-Based Methods
  for Natural Language Processing (TextGraphs-15)}, pages 67--82, Mexico City,
  Mexico. Association for Computational Linguistics.

\bibitem[{Wang et~al.(2021{\natexlab{c}})Wang, Wang, Dang, Liu, and
  Liu}]{Wang2020ACS}
Yu~Wang, Yuelin Wang, Kai Dang, Jie Liu, and Zhuo Liu. 2021{\natexlab{c}}.
\newblock \href {https://doi.org/10.1145/3474840} {A comprehensive survey of
  grammatical error correction}.
\newblock \emph{ACM Trans. Intell. Syst. Technol.}, 12(5).

\bibitem[{Waseem and Hovy(2016)}]{waseem2016hateful}
Zeerak Waseem and Dirk Hovy. 2016.
\newblock Hateful symbols or hateful people? predictive features for hate
  speech detection on twitter.
\newblock In \emph{Proceedings of the NAACL student research workshop}, pages
  88--93.

\bibitem[{Wei et~al.(2022{\natexlab{a}})Wei, Bosma, Zhao, Guu, Yu, Lester, Du,
  Dai, and Le}]{wei2022finetuned}
Jason Wei, Maarten Bosma, Vincent Zhao, Kelvin Guu, Adams~Wei Yu, Brian Lester,
  Nan Du, Andrew~M. Dai, and Quoc~V Le. 2022{\natexlab{a}}.
\newblock \href {https://openreview.net/forum?id=gEZrGCozdqR} {Finetuned
  language models are zero-shot learners}.
\newblock In \emph{International Conference on Learning Representations}.

\bibitem[{Wei et~al.(2022{\natexlab{b}})Wei, Tay, Bommasani, Raffel, Zoph,
  Borgeaud, Yogatama, Bosma, Zhou, Metzler et~al.}]{emergent_abilities}
Jason Wei, Yi~Tay, Rishi Bommasani, Colin Raffel, Barret Zoph, Sebastian
  Borgeaud, Dani Yogatama, Maarten Bosma, Denny Zhou, Donald Metzler, et~al.
  2022{\natexlab{b}}.
\newblock Emergent abilities of large language models.
\newblock \emph{TMLR}.

\bibitem[{Weidinger et~al.(2022)Weidinger, Uesato, Rauh, Griffin, Huang,
  Mellor, Glaese, Cheng, Balle, Kasirzadeh, Biles, Brown, Kenton, Hawkins,
  Stepleton, Birhane, Hendricks, Rimell, Isaac, Haas, Legassick, Irving, and
  Gabriel}]{weidinger2022taxonomy}
Laura Weidinger, Jonathan Uesato, Maribeth Rauh, Conor Griffin, Po-Sen Huang,
  John Mellor, Amelia Glaese, Myra Cheng, Borja Balle, Atoosa Kasirzadeh,
  Courtney Biles, Sasha Brown, Zac Kenton, Will Hawkins, Tom Stepleton, Abeba
  Birhane, Lisa~Anne Hendricks, Laura Rimell, William Isaac, Julia Haas, Sean
  Legassick, Geoffrey Irving, and Iason Gabriel. 2022.
\newblock \href {https://doi.org/10.1145/3531146.3533088} {Taxonomy of risks
  posed by language models}.
\newblock In \emph{2022 ACM Conference on Fairness, Accountability, and
  Transparency}, FAccT '22, page 214–229, New York, NY, USA. Association for
  Computing Machinery.

\bibitem[{Wiegand et~al.(2019)Wiegand, Ruppenhofer, and
  Kleinbauer}]{wiegand2019detection}
Michael Wiegand, Josef Ruppenhofer, and Thomas Kleinbauer. 2019.
\newblock Detection of abusive language: the problem of biased datasets.
\newblock In \emph{Proceedings of the 2019 conference of the North American
  Chapter of the Association for Computational Linguistics: human language
  technologies, volume 1 (long and short papers)}, pages 602--608.

\bibitem[{Wiegand et~al.(2018)Wiegand, Siegel, and
  Ruppenhofer}]{wiegand2018overview}
Michael Wiegand, Melanie Siegel, and Josef Ruppenhofer. 2018.
\newblock Overview of the germeval 2018 shared task on the identification of
  offensive language.

\bibitem[{Winata et~al.(2021)Winata, Madotto, Lin, Liu, Yosinski, and
  Fung}]{winata-etal-2021-language}
Genta~Indra Winata, Andrea Madotto, Zhaojiang Lin, Rosanne Liu, Jason Yosinski,
  and Pascale Fung. 2021.
\newblock \href {https://doi.org/10.18653/v1/2021.mrl-1.1} {Language models are
  few-shot multilingual learners}.
\newblock In \emph{Proceedings of the 1st Workshop on Multilingual
  Representation Learning}, pages 1--15, Punta Cana, Dominican Republic.
  Association for Computational Linguistics.

\bibitem[{Wiseman et~al.(2018)Wiseman, Shieber, and
  Rush}]{wiseman-etal-2018-learning}
Sam Wiseman, Stuart Shieber, and Alexander Rush. 2018.
\newblock \href {https://doi.org/10.18653/v1/D18-1356} {Learning neural
  templates for text generation}.
\newblock In \emph{Proceedings of the 2018 Conference on Empirical Methods in
  Natural Language Processing}, pages 3174--3187, Brussels, Belgium.
  Association for Computational Linguistics.

\bibitem[{Wolf et~al.(2017)Wolf, Miller, and Grodzinsky}]{wolf2017we}
Marty~J Wolf, Keith~W Miller, and Frances~S Grodzinsky. 2017.
\newblock Why we should have seen that coming: comments on {M}icrosoft’s
  {Tay} ``experiment,'' and wider implications.
\newblock \emph{The ORBIT Journal}, 1(2):1--12.

\bibitem[{Wolf et~al.(2020)Wolf, Debut, Sanh, Chaumond, Delangue, Moi, Cistac,
  Rault, Louf, Funtowicz, Davison, Shleifer, von Platen, Ma, Jernite, Plu, Xu,
  Le~Scao, Gugger, Drame, Lhoest, and Rush}]{Wolf2019HuggingFacesTS}
Thomas Wolf, Lysandre Debut, Victor Sanh, Julien Chaumond, Clement Delangue,
  Anthony Moi, Pierric Cistac, Tim Rault, Remi Louf, Morgan Funtowicz, Joe
  Davison, Sam Shleifer, Patrick von Platen, Clara Ma, Yacine Jernite, Julien
  Plu, Canwen Xu, Teven Le~Scao, Sylvain Gugger, Mariama Drame, Quentin Lhoest,
  and Alexander Rush. 2020.
\newblock \href {https://doi.org/10.18653/v1/2020.emnlp-demos.6} {Transformers:
  State-of-the-art natural language processing}.
\newblock In \emph{Proceedings of the 2020 Conference on Empirical Methods in
  Natural Language Processing: System Demonstrations}, pages 38--45, Online.
  Association for Computational Linguistics.

\bibitem[{Xenos et~al.(2021)Xenos, Pavlopoulos, and
  Androutsopoulos}]{xenos2021toxicity}
Alexandros Xenos, John Pavlopoulos, and Ion Androutsopoulos. 2021.
\newblock \href {https://doi.org/10.18653/v1/2021.woah-1.15} {Context
  sensitivity estimation in toxicity detection}.
\newblock In \emph{Proceedings of the 5th Workshop on Online Abuse and Harms
  (WOAH 2021)}, pages 140--145, Online. Association for Computational
  Linguistics.

\bibitem[{Xia et~al.(2020)Xia, Field, and Tsvetkov}]{xia-etal-2020-demoting}
Mengzhou Xia, Anjalie Field, and Yulia Tsvetkov. 2020.
\newblock \href {https://doi.org/10.18653/v1/2020.socialnlp-1.2} {Demoting
  racial bias in hate speech detection}.
\newblock In \emph{Proceedings of the Eighth International Workshop on Natural
  Language Processing for Social Media}, pages 7--14, Online. Association for
  Computational Linguistics.

\bibitem[{Xiang et~al.(2012)Xiang, Fan, Wang, Hong, and
  Rose}]{xiang2012detecting}
Guang Xiang, Bin Fan, Ling Wang, Jason Hong, and Carolyn Rose. 2012.
\newblock Detecting offensive tweets via topical feature discovery over a large
  scale twitter corpus.
\newblock In \emph{Proceedings of the 21st ACM international conference on
  Information and knowledge management}, pages 1980--1984.

\bibitem[{Xiang et~al.(2021)Xiang, MacAvaney, Yang, and
  Goharian}]{xiang-etal-2021-toxccin}
Tong Xiang, Sean MacAvaney, Eugene Yang, and Nazli Goharian. 2021.
\newblock \href {https://aclanthology.org/2021.wassa-1.1} {{T}ox{CCI}n: Toxic
  content classification with interpretability}.
\newblock In \emph{Proceedings of the Eleventh Workshop on Computational
  Approaches to Subjectivity, Sentiment and Social Media Analysis}, pages
  1--12, Online. Association for Computational Linguistics.

\bibitem[{Xu(2020)}]{openaifake}
Adrian~Yijie Xu. 2020.
\newblock Generating fake news with openai's language models.

\bibitem[{Xu et~al.(2021)Xu, Pathak, Wallace, Gururangan, Sap, and
  Klein}]{xu-etal-2021-detoxifying}
Albert Xu, Eshaan Pathak, Eric Wallace, Suchin Gururangan, Maarten Sap, and Dan
  Klein. 2021.
\newblock \href {https://doi.org/10.18653/v1/2021.naacl-main.190} {Detoxifying
  language models risks marginalizing minority voices}.
\newblock In \emph{Proceedings of the 2021 Conference of the North American
  Chapter of the Association for Computational Linguistics: Human Language
  Technologies}, pages 2390--2397, Online. Association for Computational
  Linguistics.

\bibitem[{Xu et~al.(2020)Xu, Ju, Li, Boureau, Weston, and
  Dinan}]{xu2020recipes}
Jing Xu, Da~Ju, Margaret Li, Y-Lan Boureau, Jason Weston, and Emily Dinan.
  2020.
\newblock Recipes for safety in open-domain chatbots.
\newblock \emph{arXiv preprint arXiv:2010.07079}.

\bibitem[{Xu et~al.(2012)Xu, Jun, Zhu, and Bellmore}]{xu2012learning}
Jun-Ming Xu, Kwang-Sung Jun, Xiaojin Zhu, and Amy Bellmore. 2012.
\newblock Learning from bullying traces in social media.
\newblock In \emph{Proceedings of the 2012 conference of the North American
  chapter of the association for computational linguistics: Human language
  technologies}, pages 656--666.

\bibitem[{Yang and Klein(2021)}]{yang2021fudge}
Kevin Yang and Dan Klein. 2021.
\newblock {FUDGE}: Controlled text generation with future discriminators.
\newblock In \emph{Proc. NAACL}.

\bibitem[{Yu et~al.(2022)Yu, Zhu, Li, Hu, Wang, Ji, and
  Jiang}]{10.1145/3512467}
Wenhao Yu, Chenguang Zhu, Zaitang Li, Zhiting Hu, Qingyun Wang, Heng Ji, and
  Meng Jiang. 2022.
\newblock \href {https://doi.org/10.1145/3512467} {A survey of
  knowledge-enhanced text generation}.
\newblock \emph{ACM Comput. Surv.}
\newblock Just Accepted.

\bibitem[{Zellers et~al.(2020)Zellers, Holtzman, Rashkin, Bisk, Farhadi,
  Roesner, and Choi}]{zellers2019defending}
Rowan Zellers, Ari Holtzman, Hannah Rashkin, Yonatan Bisk, Ali Farhadi,
  Franziska Roesner, and Yejin Choi. 2020.
\newblock Defending against neural fake news.
\newblock \emph{Neurips}.

\bibitem[{Zeng et~al.(2023)Zeng, Liu, Du, Wang, Lai, Ding, Yang, Xu, Zheng,
  Xia, Tam, Ma, Xue, Zhai, Chen, Zhang, Dong, and Tang}]{Zeng2022GLM130BAO}
Aohan Zeng, Xiao Liu, Zhengxiao Du, Zihan Wang, Hanyu Lai, Ming Ding, Zhuoyi
  Yang, Yifan Xu, Wendi Zheng, Xiao Xia, Weng~Lam Tam, Zixuan Ma, Yufei Xue,
  Jidong Zhai, Wenguang Chen, P.~Zhang, Yuxiao Dong, and Jie Tang. 2023.
\newblock Glm-130b: An open bilingual pre-trained model.
\newblock \emph{ICLR}.

\bibitem[{Zhang et~al.(2022{\natexlab{a}})Zhang, Song, Li, Zhou, and
  Song}]{CTGsurvey}
Hanqing Zhang, Haolin Song, Shaoyu Li, Ming Zhou, and Dawei Song.
  2022{\natexlab{a}}.
\newblock \href {https://doi.org/10.48550/ARXIV.2201.05337} {A survey of
  controllable text generation using transformer-based pre-trained language
  models}.
\newblock arXiv.

\bibitem[{Zhang et~al.(2020{\natexlab{a}})Zhang, Liu, Xiong, and
  Liu}]{zhang-etal-2020-grounded}
Houyu Zhang, Zhenghao Liu, Chenyan Xiong, and Zhiyuan Liu. 2020{\natexlab{a}}.
\newblock \href {https://doi.org/10.18653/v1/2020.acl-main.184} {Grounded
  conversation generation as guided traverses in commonsense knowledge graphs}.
\newblock In \emph{Proceedings of the 58th Annual Meeting of the Association
  for Computational Linguistics}, pages 2031--2043, Online. Association for
  Computational Linguistics.

\bibitem[{Zhang et~al.(2020{\natexlab{b}})Zhang, Zhao, Saleh, and
  Liu}]{zhang2020pegasus}
Jingqing Zhang, Yao Zhao, Mohammad Saleh, and Peter Liu. 2020{\natexlab{b}}.
\newblock Pegasus: Pre-training with extracted gap-sentences for abstractive
  summarization.
\newblock In \emph{International Conference on Machine Learning}, pages
  11328--11339. PMLR.

\bibitem[{Zhang et~al.(2022{\natexlab{b}})Zhang, Roller, Goyal, Artetxe, Chen,
  Chen, Dewan, Diab, Li, Lin, Mihaylov, Ott, Shleifer, Shuster, Simig, Koura,
  Sridhar, Wang, and Zettlemoyer}]{https://doi.org/10.48550/arxiv.2205.01068}
Susan Zhang, Stephen Roller, Naman Goyal, Mikel Artetxe, Moya Chen, Shuohui
  Chen, Christopher Dewan, Mona Diab, Xian Li, Xi~Victoria Lin, Todor Mihaylov,
  Myle Ott, Sam Shleifer, Kurt Shuster, Daniel Simig, Punit~Singh Koura, Anjali
  Sridhar, Tianlu Wang, and Luke Zettlemoyer. 2022{\natexlab{b}}.
\newblock \href {https://arxiv.org/abs/2205.01068} {Opt: Open pre-trained
  transformer language models}.
\newblock arXiv.

\bibitem[{Zhao et~al.(2017)Zhao, Wang, Yatskar, Ordonez, and
  Chang}]{zhao2017men}
Jieyu Zhao, Tianlu Wang, Mark Yatskar, Vicente Ordonez, and Kai-Wei Chang.
  2017.
\newblock Men also like shopping: Reducing gender bias amplification using
  corpus-level constraints.
\newblock In \emph{Proc. of EMNLP}, pages 2979--2989.

\bibitem[{Zhao et~al.(2020)Zhao, Cohen, and Webber}]{zhao2020reducing}
Zheng Zhao, Shay~B. Cohen, and Bonnie Webber. 2020.
\newblock \href {https://doi.org/10.18653/v1/2020.findings-emnlp.203} {Reducing
  quantity hallucinations in abstractive summarization}.
\newblock In \emph{Findings of the Association for Computational Linguistics:
  EMNLP 2020}, pages 2237--2249, Online. Association for Computational
  Linguistics.

\bibitem[{Zhou and Zafarani(2020)}]{Zhou2020}
Xinyi Zhou and Reza Zafarani. 2020.
\newblock \href {https://doi.org/10.1145/3395046} {A survey of fake news:
  Fundamental theories, detection methods, and opportunities}.
\newblock \emph{ACM Comput. Surv.}, 53(5).

\bibitem[{Zhu et~al.(2021)Zhu, Hinthorn, Xu, Zeng, Zeng, Huang, and
  Jiang}]{zhu-etal-2021-enhancing}
Chenguang Zhu, William Hinthorn, Ruochen Xu, Qingkai Zeng, Michael Zeng,
  Xuedong Huang, and Meng Jiang. 2021.
\newblock \href {https://doi.org/10.18653/v1/2021.naacl-main.58} {Enhancing
  factual consistency of abstractive summarization}.
\newblock In \emph{Proceedings of the 2021 Conference of the North American
  Chapter of the Association for Computational Linguistics: Human Language
  Technologies}, pages 718--733, Online. Association for Computational
  Linguistics.

\bibitem[{Zhu et~al.(2022)Zhu, Xu, Ren, Lin, Jiang, and
  Yu}]{zhu-etal-2022-knowledge}
Chenguang Zhu, Yichong Xu, Xiang Ren, Bill~Yuchen Lin, Meng Jiang, and Wenhao
  Yu. 2022.
\newblock \href {https://doi.org/10.18653/v1/2022.acl-tutorials.3}
  {Knowledge-augmented methods for natural language processing}.
\newblock In \emph{Proceedings of the 60th Annual Meeting of the Association
  for Computational Linguistics: Tutorial Abstracts}, pages 12--20, Dublin,
  Ireland. Association for Computational Linguistics.

\bibitem[{Zhuo et~al.(2023)Zhuo, Huang, Chen, and Xing}]{Zhuo2023ExploringAE}
Terry~Yue Zhuo, Yujin Huang, Chunyang Chen, and Zhenchang Xing. 2023.
\newblock Exploring ai ethics of chatgpt: A diagnostic analysis.
\newblock \emph{ArXiv}, abs/2301.12867.

\end{thebibliography}
\bibliographystyle{acl_natbib}

\appendix

\section{Background: More Details}
\label{sec:app-back}
Since we focus on language generation, we use the term \textit{language models} (LMs) to refer to their classic definition as generative models (or decoders), which predict the next token given the preceding generated context. 
For the purposes of this survey, 
this paradigm also subsumes conditional (or sequence-to-sequence) LMs conditioned on inputs from different modalities such as text, image, or speech via an encoder. 
\footnote{While many different strategies to (pre-)train encoder LMs have been introduced in the literature \cite{devlin2018bert, peters-etal-2018-deep}, they are generally not conducive to generating text and are out of scope in this survey.} Unless otherwise specified, we assume that (1) the LM decoder is parameterized by a transformer architecture~\citep{NIPS2017_3f5ee243}, and (2) the LM is first pretrained on a large amount of text (ranging from 100-billions to trillions of tokens), which, together with their large number of parameters, have earned such models the name large language models.\footnote{While some of the studies we will discuss do not rely on pretraining, we highlight it here since it is one of the primary drivers of recent advances in language generation (and its associated risks)}. 
After pretraining, LMs are either used in a zero- or few-shot manner \cite{NEURIPS2020_1457c0d6}, or modified for specific tasks via finetuning all or some of
their parameters~\citep{https://doi.org/10.48550/arxiv.2107.13586}. 

The generation tasks this survey focuses on can be broadly categorized as either (1) transformation tasks, where a given input 
is transformed into a textual output such as machine translation, abstractive summarization, data-to-text generation, and stylistic re-writing, among others \citep{prabhumoye2018style,JMLR:v21:20-074, zhang2020pegasus, aghajanyan2022htlm}, (2) or open-ended tasks such as dialogue generation, prompt-based autocompletion, story generation, and more \citep{adiwardana2020towards,
guan2020knowledge}. 




\end{document}